\documentclass{article}

\usepackage{PRIMEarxiv}

\usepackage[utf8]{inputenc} %
\usepackage[T1]{fontenc}    %
\usepackage{hyperref}       %
\usepackage{url}            %
\usepackage{booktabs}       %
\usepackage{amsfonts}       %
\usepackage{nicefrac}       %
\usepackage{microtype}      %
\usepackage{lipsum}
\usepackage{fancyhdr}       %
\usepackage{graphicx}       %
\graphicspath{{media/}}     %

\usepackage{amsmath}
\usepackage{multirow}
\usepackage{makecell}

\usepackage[dvipsnames]{xcolor}
\usepackage{tikz}
\usepackage{graphicx}
\usepackage{subcaption}
\usepackage{adjustbox}
\usepackage{makecell}
\usepackage{wrapfig}
\usepackage{caption}
\usepackage{bbm}
\usepackage{dsfont}
\usepackage{multirow}
\usepackage{wrapfig}
\usepackage{todonotes}

\AtBeginDocument{%
  }

\usepackage[numbers]{natbib}

\title{From Universal to Individualized Actionability: Revisiting Personalization in Algorithmic Recourse
}

\author{
Lena Marie Budde$^{1}$ \quad
Ayan Majumdar$^{1,2}$ \quad
Richard Uth$^{3}$ \quad
Markus Langer$^{3}$ \quad
Isabel Valera$^{1}$ \\
\\
$^{1}$Saarland Informatics Campus, Saarland University, Saarbr\"ucken, Germany \\
$^{2}$Max Planck Institute for Software Systems, Saarbr\"ucken, Germany \\
$^{3}$Institute for Psychology, University of Freiburg, Freiburg, Germany \\
\texttt{s8lebudd@stud.uni-saarland.de} \quad
\texttt{ayanm@mpi-sws.org} \quad
\texttt{\{ruth, mlanger\}@uni-freiburg.de} \\
\texttt{ivalera@cs.uni-saarland.de}
}

\begin{document}
\maketitle

\begin{abstract}

Algorithmic recourse aims to provide actionable recommendations that enable individuals to change unfavorable model outcomes, and prior work has extensively studied properties such as efficiency, robustness, and fairness. However, the role of \emph{personalization} in recourse remains largely implicit and underexplored. While existing approaches incorporate elements of personalization through user interactions, they typically lack an explicit definition of personalization and do not systematically analyze its downstream effects on other recourse desiderata.

In this paper, we formalize personalization as \emph{individual actionability}, characterized along two dimensions: hard constraints that specify which features are individually actionable, and soft, individualized constraints that capture preferences over action values and costs. We operationalize these dimensions within the causal algorithmic recourse framework, adopting a pre-hoc user-prompting approach in which individuals express preferences via rankings or scores prior to the generation of any recourse recommendation. Through extensive empirical evaluation, we investigate how personalization interacts with key recourse desiderata, including validity, cost, and plausibility. 
Our results highlight important trade-offs: individual actionability constraints, particularly hard ones, can substantially degrade the plausibility and validity of recourse recommendations across amortized and non-amortized approaches.
Notably, we also find that incorporating individual actionability can reveal disparities in the cost and plausibility of recourse actions across socio-demographic groups. These findings underscore the need for principled definitions, careful operationalization, and rigorous evaluation of personalization in algorithmic recourse.

\end{abstract}


\section{Introduction}
\label{Intro}

As automated decision-making systems (ADS) are increasingly deployed in high-stakes domains such as hiring, education, and loan approval, a growing number of individuals face significant consequences from adverse decisions. This has intensified the demand for meaningful explanations and effective mechanisms to challenge unfavorable and often opaque algorithmic outcomes.  {For example, a bank’s black-box ADS may deny a loan application, leaving the applicant without clarity regarding the decision and, more importantly, what they could do to overturn it.} Accordingly, these concerns have become central to modern AI regulation and governance frameworks~\cite{CommissionAIAct, BletchleyDeclaration, G20, Unesco}, motivating the development of methods that go beyond explanation to support actionable responses to automated decisions.

Algorithmic recourse (AR)~\cite{10.1145/3287560.3287566} has emerged as a promising approach to address these challenges. 
Assuming a fixed, deployed ADS, it aims to provide individuals with  {\emph{minimal-effort, valid, and actionable recommendations}—feature changes that are realistically implementable, and, if applied, would lead to a favorable decision}.
For instance, in the lending example, the bank could offer the applicant a recourse recommendation, such as increasing income by 10\% and reducing the requested loan amount by 3\%, which would lead to loan approval if the application were resubmitted.

Early work in recourse focused on ensuring \emph{global actionability}, typically by excluding inappropriate features and actions (e.g., reducing age or changing race)~\cite{10.1145/3287560.3287566}.  Subsequent research demonstrated that such constraints are insufficient on their own: to ensure true actionability, recourse must also account for the underlying causal relationships among features~\cite{karimi2020algorithmic,karimi2021algorithmic}.
For instance, recommending increases in both salary and savings ignores that savings depend on salary. Hence, disregarding causality can make some suggestions suboptimal and invalid. While accounting for both global and causal actionability improves recourse in general, it does not guarantee that recommendations are feasible for specific individuals with unique circumstances.

In practice, the boundary between actionable and non-actionable recommendations is highly context-dependent. A change that is feasible for one individual may be costly, impractical, or even impossible for another. Moreover, features that are individually actionable may not be jointly attainable~\cite{Venkatasubramanian_2020}. Relying solely on global notions of actionability effectively shifts the burden of change onto individuals, potentially reframing structural constraints as personal shortcomings. As noted by~\citet{Venkatasubramanian_2020}, ``\textit{What if (as seems likely) disadvantaged people are also those who most frequently need to resort to recourse, yet have the least bandwidth—in terms of time, energy, money, and social connections—to do so?}'' When individuals cannot implement suggested changes, recourse fails to provide meaningful agency~\cite{wachter2017counterfactual}, risks unfair moral judgment, and may disproportionately harm already marginalized groups~\cite{10.1145/2783258.2783311}.

These challenges highlight the importance of developing recourse methods that are grounded in what is realistically achievable for each individual.
Accordingly, both psychological research and recent qualitative work on AR suggest that recourse recommendations should be tailored to individuals’ needs and constraints (\emph{personalized AR})~\cite{2022personalized_pref_eliction, Uth_2025, Wang_2023, Esfahani_2024}. However, the necessary conditions for what constitutes a personalized recourse recommendation, as well as how personalization interacts with other key desiderata--particularly validity, plausibility, cost, and fairness--remain largely underexplored within the AR community.

Prior work primarily focuses on personalizing cost functions~\cite{yadav2021low}, often through iterative approaches that update recommendation sets~\cite{2022personalized_pref_eliction,abrate_2024}, i.e., modeling how individuals perceive the cost of different feature changes. However, by \textit{not explicitly accounting for which features are individually actionable from the outset}, these approaches may initiate the iterative process with recommendations that are infeasible for a given individual. Furthermore, the \textit{quantitative impact} of incorporating actionability constraints on other key desiderata, as well as their broader normative implications, remains largely unexplored.

To address these gaps, we advocate for framing \textbf{personalization in terms of individualized actionability}: %
a recourse recommendation should, at a minimum, be realistically implementable within an individual’s unique constraints, ensuring that suggested actions are both feasible and achievable for them. 
For a holistic understanding of different constraints on recourse, we distinguish between \emph{hard and soft individual actionability constraints}. While hard individual actionability constraints define which features are individually actionable, i.e., what features an individual is actually able to change, soft individual actionability constraints capture preferences of an individual's perceived ease of acting on specific features compared to others. We formalize and study these constraints within causal AR, extensively examining their quantitative impact on key recourse desiderata as well as their broader implications for fairness.

Concretely, we \textbf{(1) formalize individual actionability constraints within the causal algorithmic recourse framework}, including both hard and soft constraints. To \textit{enable large-scale empirical studies}, we build on the amortized causal recourse solver, CARMA~\cite{CARMA}, to \textbf{(2) propose individualized CARMA (iCARMA), an efficient amortized solver for scalable personalized causal recourse}. Through extensive experiments on synthetic and real-world datasets, we \textbf{(3) show how incorporating individual actionability affects key recourse metrics}. Our results reveal that (i) hard constraints conflict with universal attainability of recourse, and (ii) soft constraints introduce less severe trade-offs but do not prevent recommendations involving the \textit{least-preferred} features. Finally, through a \textbf{(4) fairness case study}, we \textbf{demonstrate that ignoring individual actionability aspects can lead to misleading fairness assessments} of recourse systems by obscuring hidden disparities. Overall, our findings highlight that global notions of actionability are insufficient for recourse deployment, and that heterogeneous individual constraints—together with broader socio-technical solutions—are essential to fully support user agency.

\noindent
\textit{Paper Outline.} We begin in Section~\ref{Section_RelWork} by reviewing related work and identifying the gaps our work addresses. Section~\ref{Section_IndivAR} formalizes individual actionability constraints in causal recourse, and Section~\ref{section_AmortizedAR} introduces iCARMA, our amortized approach for scalable personalized causal recourse. Finally, Section~\ref{sec:eval} presents empirical evaluations for quantitatively studying trade-offs and a case study examining fairness.

\section{Related Works and Their Limitations}
\label{Section_RelWork}

In \emph{Explainable AI}, algorithmic recourse aims to \emph{provide agency} to individuals adversely affected by negative decisions from a \textit{fixed and deployed} algorithmic decision-making system. A related concept is \textit{counterfactual explanations}~\cite{wachter2017counterfactual}, which identify minimal input changes that would alter a model’s prediction. However, without causal considerations, such explanations mainly support \emph{auditing} predictive systems and uncovering issues such as hidden biases, rather than providing actionable guidance.  {Relatedly, \textit{strategic classification}~\cite{strategicClassification} models a two-player game in which users \emph{manipulate features a priori} in anticipation of a classifier, while the classifier \emph{adapts} to mitigate such manipulation.} In contrast, algorithmic recourse is \emph{user-centric}: it identifies actions individuals can take~\cite{10.1145/3287560.3287566} by leveraging feature causal relations~\cite{karimi2020algorithmic,karimi2021algorithmic} to change outcomes while assuming that the \textit{predictive system remains fixed}. 

To meaningfully enhance user agency, interdisciplinary and qualitative works~\cite{Wang_2023,koh24,tomu25,Uth_2025} highlight the \emph{need to incorporate end-user preferences} in recourse solvers. While recent technical work has begun exploring personalization~\cite{2022personalized_pref_eliction}, a \textit{systematic understanding of its implications remains limited}.  We discuss these limitations below and provide an extended overview of related work in Appendix~\ref{RelWork_Appendix}.

\textbf{Lack of consideration for downstream causal effects.}
Most approaches define a user's action feasibility purely via constraints in the counterfactual space~\cite{karimi2022survey, valera2020survey, 10.1145/3287560.3287566, kanamori2020dace, pmlr-v108-karimi20a, Pawelczyk_2020, mahajan2020preservingcausalconstraintscounterfactual}, which prevents distinguishing between an \textit{action a user can directly perform} and a \textit{change that occurs only via causal propagation}. For example, there might be a significant difference in whether a user can actively increase their savings (e.g., by selling some possessions) or only increase them by obtaining a higher income. However, existing work on personalized recourse fails to account for this distinction~\cite{2022personalized_pref_eliction, Wang_2023, Esfahani_2024,Yetukuri_2023}.

\textbf{Omission of individual hard constraints.} 
While \citet{2022personalized_pref_eliction} and \citet{yadav2021low} explicitly focused on personalization, both approaches relied on eliciting individual cost functions. Other approaches~\cite{Mothilal_2020, Wang_2021,abrate_2024} leveraged giving users multiple counterfactuals for selection. However, none of these works ensure that proposed recourse recommendations do not include individually non-actionable features, potentially leading to intractable solutions.

\textbf{Limited quantitative understanding of individual constraints.}  
While many recourse methods can \emph{mathematically} encode user-level constraints \cite{10.1145/3287560.3287566, 10.1145/3287560.3287569, pmlr-v108-karimi20a, 10.1145/3375627.3375850,dandl2020multi, Wang_2023_flexible, cheon2024featureresponsivenessscoresmodelagnostic}, prior work rarely examines how these constraints interact with key recourse objectives. Even studies that focus on personalization have limitations: some are primarily qualitative, emphasizing user experience and interface design \cite{Wang_2023,Esfahani_2024}, while others introduce preference-aware methods without analyzing the quantitative trade-offs arising from individual constraints \cite{2022personalized_pref_eliction}. Hence, the \textit{precise effects of individual constraints remain poorly understood.} Furthermore, fairness analyses typically consider group-level recourse cost differences \cite{rawal2020beyond, bell2024fairness, li2025fair}, but individuals also differ in their \emph{capacity to execute} actions. Unequal access to resources may render certain actions infeasible for specific groups, an aspect invisible to frameworks that ignore individual actionability.

\textbf{Limited normative discussion.}  
Existing work on personalized recourse rarely discusses \emph{design choices} behind \emph{incorporating individual constraints in the optimization problem}. In particular, prior studies rarely consider the normative implications of how user preferences are collected or how they are represented and encoded for optimization.
In contrast, our work systematically examines the effects hard as well as soft individual actionability constraints, highlighting aspects for normative discussion in future interdisciplinary works.

\section{Individual Actionability in Algorithmic Recourse}
\label{Section_IndivAR}

Given that \emph{accounting for causal relationships between features} is essential for actionable recourse~\cite{karimi2020algorithmic}, we focus on studying how \emph{individual actionability constraints} affect key objectives within a causality-aware framework.  {This section provides background on causal AR and its key desiderata (Section~\ref{Background_CausalAR})}, formalizes individual actionability constraints within causal AR (Section~\ref{SubSection_IndivConst}), and describes the \emph{minimal prompting scheme} for eliciting these constraints (Section~\ref{sec:prompting}).

\subsection{ {Background: Causality for Actionable Recourse}}
\label{Background_CausalAR}

Let $u$ be an individual affected by a deployed, fixed, and binary automated decision-making system (ADS) $h$, and let $x_u$ denote the factual input features provided to the ADS. If the ADS predicts negatively $h(x_u)=0$, $u$ seeks \textit{actionable recommendations} to modify their features, \emph{improve their profile}, and overturn the decision through AR~\cite{10.1145/3287560.3287566}.

Prior work~\cite{karimi2020algorithmic,karimi2021algorithmic} showed that ensuring recourse recommendations are \emph{actionable and optimal} requires accounting for the \emph{causal relationships} among features.
Within the causal framework~\cite{pearl2009causality2}, these relationships are represented using a structural causal model (SCM), where features $X$ are linked through structural functions $F$. 
\\
\\Recourse actions are modeled as \textit{causal interventions} on the SCM. Given an action $A:= do(\{x_{u,i}:= a_i\}_{i \in I})$, where $I$ denotes the subset of features acted upon, the modified structural functions $F_A$ are obtained by replacing the $i^{th}$ function with $X_i:= a_i$ for all $i \in I$. The recourse counterfactual $x_u^{CF}$—i.e., the features $u$ would attain after the action—is given by $F_A(F^{-1}(x_u))$. 
We provide a more detailed discussion on causality and recourse in Appendix~\ref{SCM_Appendix}.

To produce \textit{optimal} recommendations, AR \emph{primarily} seeks actions that are (i) \textbf{valid}, meaning they overturn the original ADS prediction $h(x_u^{CF}) \neq h(x_u)$ without which AR fails its purpose, and (ii) \textbf{minimum cost}, ensuring the required changes impose \emph{minimal effort} on the individual. Existing literature~\cite{karimi2020algorithmic,von2022fairness,CARMA} has considered \emph{cost functions}, e.g., $\ell_p$-norms, as a proxy for \emph{user effort}. In addition, existing recourse methods aim to satisfy (iii) \textbf{feasibility}, requiring actions to lie in the set $\mathcal{F}$ of \emph{realistically implementable} actions, and (iv) \textbf{plausibility}, requiring the resulting counterfactual to lie in the \emph{distributionally plausible set} $\mathcal{P}$, acting as a proxy for ensuring \emph{realistic feature profiles}~\cite{heidari2019long} and providing in-distribution profiles essential for reliability in algorithmic pipelines. Hence, AR solves the following optimization problem~\cite{karimi2021algorithmic}:

\begin{equation}
\text{arg} \min_{A} {\text{ cost}(A; x_u)}, \,\text{such that } h(x_u^{CF}) \neq h(x_u)\, \,\text{and } 
    \, {A \in} \mathcal{F} 
    \, , 
    x_u^{CF} \in \mathcal{P} 
\end{equation}

 {These considerations and mechanisms distinguish AR from \emph{counterfactual explanations}~\cite{wachter2017counterfactual}, which, without causal modeling, are primarily useful for auditing or debugging deployed ADS rather than providing actionable recommendations. Importantly, without causal considerations, recourse would assume each feature to be \textit{independently manipulable}, which has been theoretically shown to be \textit{suboptimal}~\cite{karimi2020algorithmic} regarding cost, validity, and also the number of features required of users to change~\cite{CARMA}. AR's mechanism also differentiates it from \textit{strategic classification}~\cite{strategicClassification}, where individuals manipulate their features \emph{before} receiving predictions and the classifier $h(\cdot)$ is \emph{jointly} modified to mitigate such gaming.}

\subsection{{Individual Actionability Constraints in Recourse}}
\label{SubSection_IndivConst}

 {
While causal consideration enhances recourse actionability, incorporating individual actionability constraints—capturing users' preferences and capabilities over different features—is a critical requirement for AR systems. Yet, their \textit{systematic impact on recourse generation}, particularly in causality-aware methods, remains largely unexplored. Individual constraints may interact non-trivially with the causal structure: SCMs allow recourse methods to account for downstream effects, often producing sparser and lower-cost recommendations~\cite{CARMA} when features high in the causal order are used. However, \textit{user preferences over which features are harder to change may align or conflict with this ordering}. For example, a feature influencing many downstream variables may be especially difficult for a user to modify, emphasizing the need to study the effect of individual actionability on causal AR.
}

To formalize the study of individual constraints in causal AR, we define the following individualized version of AR~\cite{karimi2021algorithmic}, with the individual-specific modifications highlighted in red:
\begin{equation}
    \text{arg} \min_{A} \textcolor{red}{\text{ cost}_u(A; x_u)} 
    \, \text{so that } h(x_u^{CF}) \neq h(x_u), \,\text{and}
    \, \textcolor{red}{A \in \mathcal{F}_u} \subseteq \mathcal{F},
    \,
    \, x_u^{CF} \in \mathcal{P}
    \label{Individual_Actionability}
\end{equation}
$\mathcal{F,P}$ denotes the global feasibility and plausibility sets as defined by existing work in AR. The individualized cost function $\text{cost}_u(A; x_u)$ represents a proxy for the effort required by user \(u\) to perform an action \(A\). \(\mathcal{F}_u\) denotes the set of \emph{individually feasible} actions, i.e., the actions that user \(u\) can realistically execute \emph{based on individual actionability constraints}.
We define this set concretely as $\mathcal{F}_u = \{i, x_i\in[a,b]\}_{i \in AF_u}$, where $AF_u$ denotes the set of features individually actionable for $u$ (cardinality $k_u$) and $AF_u\subset AF$, i.e., the features a user can act on are a subset of globally actionable features $AF$ (cardinality $k$).

Hence, \textit{two distinct types of constraints} define individual actionability: (i) hard constraints, which specify which features are individually actionable, and (ii) soft, individualized constraints, which capture preferences over action values and associated costs across different features.

\noindent
{\textbf{Hard Constraints.}} 
A user may be \textit{unable to take any action} on a given feature, either because the feature is globally non-actionable or because the user faces circumstances that render a \textit{globally actionable} feature effectively non-actionable. We refer to such features as \textit{individually non-actionable} and assume them to be determined by the hard constraints in the optimization problem above.  {Thus, such features are strictly not available for recourse generation.} In the lending use case, for example, we have previously noted that neither income nor loan amount should be assumed actionable for all individuals. 
In contrast, features on which a user can perform actions are referred to as \textit{individually actionable features}.

\noindent
\textbf{Soft Constraints.}
Even among individual actionable features, the effort required to change a feature by a particular margin varies from feature to feature, and even the margin by which a feature can be changed is, in general, not equal among features.  {Assume someone who can change the income easily but at a maximum of $10\%$ might be able to change savings by $50\%$, albeit finding that very hard. \emph{These very nuanced preferences are subsumed under soft constraints}. 
Beyond the feasibility set, soft constraints also determine the individualized cost function.}
For the sake of simplicity, we assume the individual cost function to be of the form
$\text{cost}_u(A; x) = \sum_{i \in AF_u} w_{u,i} \cdot cost_u(x_{i})$,
where $w_{u,i}$ is the \textbf{actionability weight}, capturing the difficulty of acting on a particular feature.

\subsection{Prompting Users for Their Individual Actionability Constraints}
\label{sec:prompting}
Having formalized individual actionability constraints in the context of causal AR, we now discuss how this information can be obtained from users in practice. Since such constraints and user-level cost parameters are not directly observable, they must be elicited from the affected individual.

Recent psychological~\cite{Langer_2024, Uth_2025} and qualitative studies~\cite{Wang_2023} suggest that eliciting preference rankings is a low-effort way to improve user satisfaction and trust when interacting with decision-support systems. Such approaches enable users to express preferences without requiring mathematical knowledge to specify full cost functions for the recourse objective~\cite{koh24}. 
In this work, we consider a \textbf{minimal \textit{pre-hoc} prompting scheme} inspired by those studies that captures essential information about individual actionability before solving the recourse optimization problem. Specifically, we explore a flexible scheme allowing for expressing hard constraints, preference ranks, and scoring over actionable features. We consider three prompting schemes of increasing granularity: (i) users group globally actionable features into individually actionable and non-actionable; (ii) users additionally rank actionable features by difficulty, providing indirect information about individualized costs; and (iii) users assign Likert-scale difficulty scores to each feature while marking features as non-actionable. Note that, in the latter case, the number of scale levels should be calibrated to the number of globally actionable features.

\textbf{Faithfulness of collected individual constraints.}
The proposed prompting scheme is capable of capturing the two key aspects of individual actionability that arise from a user's personal circumstances. Depending on their socioeconomic situation, users may (i) identify features that are entirely non-actionable and (ii) perceive actionable features as requiring different levels of effort. 
Because these aspects reflect users' existing capabilities and constraints, they can be meaningfully elicited \textit{pre-hoc} and incorporated directly into the AR optimization problem (Eq.~\ref{Individual_Actionability}). We further assume that users report such preferences truthfully. This assumption is reasonable because AR is \textit{not strategic}: misreporting preferences does not provide systematic advantages to users. Since recourse recommendations must ultimately be implemented in practice, falsely reported preferences would likely produce recommendations that are misaligned with users' actual capabilities, making them suboptimal or infeasible.

 \textbf{Pre-hoc vs. post-hoc prompting.} We study personalization effects that are rooted in the optimization problem and occur independently of a specific prompting scheme. We operationalize our analysis through pre-hoc prompting because of its minimal nature.
 In contrast, existing work~\cite{Wang_2023,2022personalized_pref_eliction,Esfahani_2024} has primarily explored \emph{post-hoc} elicitation, in which users provide feedback after receiving candidate recourse recommendations. While interactive feedback can help refine recommendations,
it can not reliably prevent the recourse generation system from producing infeasible recommendations from the outset.
 Furthermore, prolonged iterative interactions may increase cognitive load for users and raise practical concerns related to the privacy of deployed systems~\cite{pawelczyk2022privacyrisksalgorithmicrecourse}.
 However, in practice, the pre-hoc approach might well be complemented by post-hoc prompting, for example, by collecting initial hard constraints and preference rankings prior to generating recourse recommendations and subsequently refining feasibility constraints through interaction. 
While such informative prompting can capture \textit{additional individual constraints} for the AR optimization problem, they can potentially lead to more significant trade-offs, which must be studied in future work.

\section{Amortized Causal Algorithmic Recourse with Individual Actionability Constraints}
\label{section_AmortizedAR}

\begin{figure}[t]
    \centering
    \includegraphics[width=0.85\linewidth]{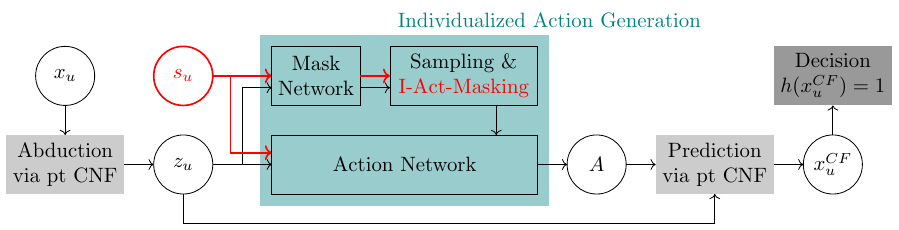}
    \caption{iCARMA architecture (adapted from~\cite{CARMA}); red parts indicate modifications to enable \textbf{individualized} recommendations.}
    \label{fig: iCARMA}
\end{figure}

In this section, we propose an efficient solver for the individualized causal recourse problem introduced in Section \ref{Section_IndivAR}, enabling large-scale quantitative analysis of personalization effects. Brute-force approaches to the \emph{individualized causal recourse problem} in Eq. \eqref{Individual_Actionability} are computationally infeasible at scale, and no existing efficient solver is both causality-aware and capable of handling hard as well as soft individual actionability constraints (see Section \ref{Section_RelWork}). To address this gap, we extend the amortized causal algorithmic recourse framework \textbf{CARMA}~\cite{CARMA} to incorporate individual constraints, which we assume are elicited through the minimal prompting scheme described in Section \ref{sec:prompting}. We refer to this extended framework as \textbf{individualized CARMA (iCARMA)}. iCARMA enables systematic studies of personalization effects at scale and allows us to examine whether amortized recourse methods remain compatible with personalization when individual actionability constraints are taken into account. 
\\
\\Figure~\ref{fig: iCARMA} illustrates the iCARMA framework, with blocks highlighted in red indicating the modifications introduced to the original CARMA to account for individual actionability.

\subsection{Background: Amortized Causal Recourse with CARMA}
\label{Background_CARMA}

Instead of solving complex combinatorial recourse problems separately for each user, CARMA~\cite{CARMA} achieves efficiency by training a neural network-based pipeline to generate recourse recommendations, i.e., predicting which feature to act upon and the required (minimal) amount of change. Moreover, CARMA enables amortization while \textit{operating in line with our causal formulation of the recourse problem}. 

CARMA uses a pre-trained causal generative model, specifically a causal normalizing flow (CNF)~\cite{CausalNF}. This model enables CARMA to operate even when the exact structural equations of the SCM are unknown. Given a causal graph and observational data, the CNF can approximate these equations, \textit{approximating the exogenous variables in the causal latent space} (see Appendix~\ref{CNF} for additional details). As shown in Fig.~\ref{fig: iCARMA}, the CNF is used for abduction, i.e.,  {computing the latent $z_u$ from features $x_u$}, and prediction, i.e.,  {computing counterfactual features} $x_u^{CF}$ from $z_u$ and the recommended action $A = do({x_{u,i} := a_i}_{i \in I})$, where $I$ denotes the set of globally actionable features chosen for intervention.

Recourse recommendations are generated via two jointly trained networks. The \textbf{Mask Network} learns a Bernoulli distribution over feature interventions, producing a binary mask $\mathcal{I}$ via the Gumbel trick~\cite{Gumbel}, while enforcing global actionability through a fixed actionability mask. The \textbf{Action Network} predicts intervention magnitudes $a_i$ given $\mathcal{I}$ and $z_u$. Training combines action cost minimization with a validity regularizer and a distribution-based loss for learning feature-level interventions (more details in Appendix~\ref{carma}).

\noindent
\textit{Plausibility constraints.}
Natively, CARMA \textit{does not account} for the distributional plausibility of recommended actions. However, as discussed in Section~\ref{Background_CausalAR}, plausibility is key, as it serves as a proxy for whether the profiles resulting from recourse recommendations are attainable for users. It is also essential for preventing recourse profiles from falling \textit{out of distribution}, which can degrade the performance of ML–based recourse pipelines and undermine the reliability of the deployed ADS. To address this limitation, we \emph{incorporate distributional plausibility constraints} into CARMA’s objective, thereby improving the reliability of the resulting recourse recommendations (see Appendix~\ref{iCARMA_loss_modified} for more details).

\subsection{Incorporating Actionability Constraints for iCARMA}
 Now, we discuss the incorporation of the different individual actionability constraints in CARMA to develop the individualized, amortized causal recourse solver iCARMA.
\subsubsection{Enforcing Hard Constraints}
CARMA’s architecture can be easily adapted to exclude individually non-actionable features. In iCARMA, we replace the population-level actionability mask with an \textbf{individualized actionability mask} (I-Act-Masking; Fig.~\ref{fig: iCARMA}), whose entries are $1$ iff feature $f$ is individually actionable for user $u$, and $0$ otherwise.

\subsubsection{From User Preferences to an Individualized Cost Function} \label{sec:indiv_cost}
To incorporate user preferences over features that are provided as hard and soft actionability constraints in iCARMA, we operationalize them through \textbf{actionability scores}.
Actionability scores translate preferences elicited from users through interpretable schemes, e.g., ranks or Likert-scale scores as discussed in Section~\ref{sec:prompting}, to a formal notion to be used in the optimization pipeline. Note that our optimization pipeline is \textit{independent of the specific interface and elicitation scheme} used to obtain the preferences from end-users.
For user $u$ and feature $i$, the actionability score $s_{u,i}$ can capture \textit{both hard and soft preference constraints} and
{is defined by}
\begin{align}
    s_{u,i} = \begin{cases}
        s^{max} & \text{if feature } i \text{ is }\textbf{individually non-actionable}\text{ or globally non-actionable} \\
        pref_{u}(i) & \text{else}
    \end{cases}
    \label{pref_eq}
\end{align}

\begin{wrapfigure}[13]{r}{0.3\textwidth}
\centering
 \vspace{-15pt}
\includegraphics[width=0.3\textwidth, trim={0 1.4cm 0 0}]{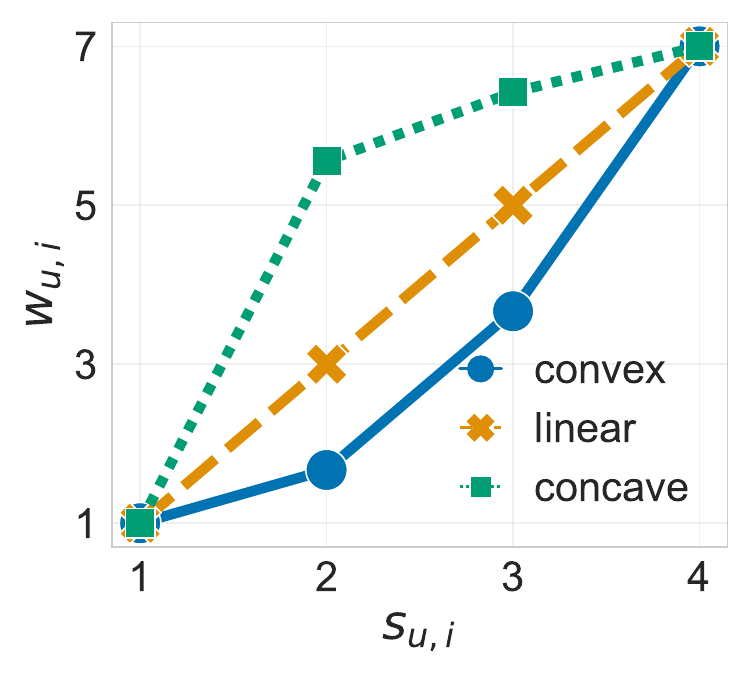}
 \vspace{3pt}
\caption{Actionability weights and cost profiles ($k=4$, $w^{\max}=7$).}
\label{fig:actionability_weights}
\end{wrapfigure}

For the general case of scoring, $s^{\max}$ equals the number of Likert Scale categories (which is to be selected on a case-by-case basis), 
and $pref_{u}(i)$ is a user-specific function mapping from the set of individually actionable features $AF_u$ to the set $\{1, \dots,s^{max}-1\}$. This function is fully determined by the Likert Scale scores given by the user.

For the special case of ranking, $s^{max} = k+1$  and $pref_{u}(i)$ is bijective, mapping to the set $\{1, \dots, k_u\}$  {, fully determined by how the user $u$ ranked their features}. For the case of binary preferences, the user only gives a classification of features as individually actionable or individually non-actionable. In this case, we also define $s^{max} = k+1$ but $pref_{u}(i) = 1$. Setting $\mathrm{pref}_u(i)$ constant indicates user indifference over the ease of acting on individually actionable features.

The actionability scores are then converted into feature-wise weights, which define the individualized cost function central to the individualized causal recourse optimization problem (Section~\ref{SubSection_IndivConst}) and are used directly by iCARMA.
\begin{equation}
\label{cost_w}
w_{u,i}= w^{max} - (w^{max}- w^{min})\cdot \left(1-\left(\frac{s_{u,i}-1}{s_u^{max}-1}\right)^{\alpha}\right)
\end{equation}
where $\alpha$ determines the shape and slope of the penalty over increasing $s_{u,i}$ and $s_{u}^{max}= \max(2, \max_{i \in AF_u}(s_{u,i}))$. For comprehensive analysis, we consider different $\alpha$ profiles for our evaluations. For $\alpha =1$, $w_{u,i}$ increases linearly and the increase in penalty is constant over all adjacent preferences (\textbf{linear} cost profile). For $\alpha > 1$, the shape is convex, i.e., the increase in actionability weight increases as $s_{u,i}$ increases, i.e., the user's preference worsens (\textbf{convex} cost profile). Conversely, for $\alpha < 1$, the shape is concave, i.e., the increase in actionability weight decreases as $s_{u,i}$ increases (\textbf{concave} cost profile). Figure \ref{fig:actionability_weights} illustrates the behavior of $w_{u,i}$ for different configurations.
Implementation details for the corresponding modifications to the CARMA optimization pipeline are provided in Appendix~\ref{iCARMA}. Specifically, actionability weights individualize the $\ell_2$-distance-based cost in CARMA's Action Network loss, while the Mask Network’s distribution-based loss is adapted via \textit{individualized actionability priors} $\pi_{u,i}$.

\subsubsection{Operational Considerations}
Ideally, iCARMA would use real-world user feedback to capture individual actionability preferences. 
To allow for a systematic study of particular edge cases in absence of real-world preference data, our implementations supports \textit{sampling actionability scores} from predefined distributions and applying a data-augmentation-style procedure during training (see Appendix~\ref{iCARMA}). 
Training-time sampling is uniform over all possible preferences
ensuring the system can handle diverse preference distributions at deployment. At test time, specific preference distributions 
are sampled to systematically probe the limits of causal recourse under individualized actionability, exploring both typical and boundary cases. However, sampling does not replace real user feedback, which is essential for deployment.

Actionability scores cannot fully capture users' ground-truth cost functions~\cite{2022personalized_pref_eliction}, so we treat the individualized cost function as a surrogate for true user effort~\cite{heidari2019long}. They encode soft constraints by penalizing changes to features with ``bad'' preference scores while remaining individually actionable, assigning higher costs to less desirable changes. We derive the feature-wise weights of the cost function from these \textit{scores and global hyperparameters}, and choose them using a validation set to perform well across a broad population. Although we consider different profiles for a more comprehensive analysis, ranking or scoring during elicitation naturally introduces information loss between intrinsic user capabilities and AR optimization. This limitation can be mitigated by additional post-hoc or interactive elicitation~\cite{2022personalized_pref_eliction, Yetukuri_2023}, but this additional step may not only increase cognitive load on the user but also  further shrink the recourse solution space, making the trade-offs in recourse metrics more apparent. We leave such \textit{complementary advances in prompting approaches} and their further impact on recourse for future work.

\section{Empirical Evaluation}
\label{sec:eval}

In the previous sections, we formalized individualized causal recourse with user-specific actionability constraints and introduced iCARMA for generating personalized causal recourse at scale. 

Here, we systematically investigate how individual constraints influence key recourse objectives and present a case study on how individual actionability affects fairness assessments in algorithmic recourse.  Our evaluation addresses the following research questions:
\begin{description}
\item[RQ1:] How do individual hard actionability constraints impact key recourse desiderata? 
\item[RQ2:] How do individual soft actionability constraints impact key recourse desiderata?
\item[RQ3:] How does individual actionability expose disparities and fairness issues in recourse systems?%
\end{description}

\subsection{Evaluation Metrics}

\noindent In line with prior work~\cite{karimi2022survey} and our discussion in Section~\ref{Background_CausalAR}, we consider the following metrics to operationalize the key desiderata of AR for understanding the quantitative trade-offs of individual actionability constraints.

\noindent
\textbf{Validity.}  This is the fundamental goal of recourse; we measure the proportion of deployment-time users $u$ who request recourse and for whom the counterfactual recommendation $x^{CF}_u$ yields {the desired} positive decision, i.e., $h(x^{CF}_u) = 1$.

\noindent
\textbf{Cost.}  As a core optimality metric, we report the population-level mean and standard deviation of the $\ell_2$ cost of feature changes. We adopt $\ell_2$ cost due to its widespread use in AR~\cite{karimi2022survey}, and leave extensions to more complex effort functions~\cite{heidari2019long} for future work. Importantly, as individualized recourse optimizes a user-specific weighted $\ell_2$ cost, comparisons across preference settings require reporting the \textit{unweighted} cost, with all actionability weights set to 1. %

\noindent\textbf{Plausibility.} 
The proportion of deployment-time users $u$ for whom the recourse counterfactual is \textit{at least as likely} under the data distribution as the factual instance, i.e., the log densities from the CNF satisfy $\log p(x^{CF}_u) \geq \log p(x_u)$. 
 As discussed in Sections~\ref{Background_CausalAR} and~\ref{Background_CARMA}, plausibility is crucial for ensuring that recourse profiles are realistic for users and remain in-distribution, supporting the reliability of amortized recourse pipelines and deployed decision-making models.

\noindent  \textbf{Preference Satisfaction.} Hard individual actionability constraints are enforced in the optimization and always satisfied, so no metric is reported for them. For soft constraints, we report \textbf{hard actions probability (hap)}. Individual actionability should minimize \emph{hard} actions, i.e., changes to the least preferred (yet still individually actionable) features for a user.  Thus, \textbf{hap} serves as a proxy for preference satisfaction in the soft constraint case, estimated as the proportion of users for whom the recourse solver \textit{recommends at least one hard action}. 

\noindent Note that, since validity, plausibility, and hap are population-level proportions, we do not report a standard deviation.

\subsection{ Setup}

\noindent  \textbf{Datasets.} We use two datasets for our analysis. First, we leverage the semi-synthetic \textit{Loan} data~\cite{karimi2020algorithmic,CARMA}, generated entirely from a predefined SCM modeling the \textit{German Credit} dataset. 
It contains seven features: Age, Gender, Education, Income, Savings, Loan Duration, and Amount, with the last four globally actionable  (further details in Appendix~\ref{loan_Functions}).
Second, we demonstrate transfer to real-world data using the \textit{GiveMeSomeCredit (GMSC)} dataset. Unlike \textit{Loan},  {\textit{GMSC} features were collected from real loan applicants~\cite{kaggle_givemesomecredit} and the true equations \textit{are unknown}.} We adopt the cost-correlation structure from~\cite{2022personalized_pref_eliction} as the underlying causal graph, treating Age and Number of Dependents as immutable and Income, Number of Real-Estate Loans, Number of Open Credit Lines, and Revolving Utilization of Unsecured Lines as globally actionable (see Appendix~\ref{apx:real-world-details} for additional details and Appendix~\ref{real_world_res} for more real-world analysis).

\noindent  \textbf{Preference Sampling.} In the absence of representative real-world data on {user preferences}, we sample actionability scores to represent particular \emph{use cases of interest} and provide a \textit{holistic} quantitative analysis. The precise preference configuration is reported for each analysis separately, and additional details are provided in Appendices ~\ref{Sampling} and ~\ref{Appendix_PrefDist_Sampling}.

\noindent  \textbf{Causal Recourse Solvers.}
For both datasets, we generate recourse recommendations using our proposed \textbf{amortized individualized solver} iCARMA. We train it under a uniform distribution over individually actionable features, meaning that both which features and how many are actionable per user are sampled uniformly, with plausibility included as a regularization term and using ten random seeds. In the absence of true equations, iCARMA is essential for studying \textit{GMSC}, since it can leverage the CNF to approximate the unknown relations. Additional details are given in Appendix~\ref{Appendix_PracticalDetails}.

For the \textit{Loan} dataset, we also generate recommendations with a \textit{Causal Oracle}~\cite{karimi2020algorithmic}. With full access to the SCM and the structural equations, it solves \emph{non-amortized} optimization problems for each user via grid search over the actionable feature space, while enforcing individual actionability preferences and plausibility as a hard constraint. This approach provides a gold-standard reference for assessing the impact of individual actionability \textit{independent of amortization}.

\clearpage

\begin{wraptable}[17]{r}{0.5\columnwidth}
\vspace{-12pt}
\centering
\caption{Impact of hard actionability constraints on recourse metrics for amortized iCARMA and non-amortized Oracle.}
\label{tab:exp_1_uniform}
\resizebox{0.5\columnwidth}{!}{%
\begin{tabular}{ccccc}
\toprule
\makecell{\textbf{dataset,} \\ \textbf{no. of individually} \\ \textbf{actionable features}} & \textbf{method} & \textbf{validity} & \textbf{plausibility} & \textbf{cost} \\
\midrule
\multirow{2}{*}{Loan, all (4)} & oracle & 0.959 & 1.0 & $0.057_{\pm 0.064}$  \\
& iCARMA & 0.999 & 0.886 & $0.069_{\pm 0.077}$ \\
\cmidrule{2-5}
\multirow{2}{*}{Loan, 3}  & oracle & 0.796 & 1.0 & $0.07_{\pm 0.085}$ \\
& iCARMA & 0.995 & 0.689  & $0.073_{\pm 0.084}$ \\
\cmidrule{2-5}
\multirow{2}{*}{Loan, 2} & oracle & 0.539 & 1.0 & $0.078_{\pm 0.096}$ \\
& iCARMA & 0.991 & 0.481  & $0.09_{\pm 0.111}$ \\
\cmidrule{2-5}
\multirow{2}{*}{Loan, 1}  & oracle & 0.22 & 1.0 & $0.077_{\pm 0.096}$ \\
& iCARMA & 0.836 & 0.258  & $0.076_{\pm 0.081}$ \\
\cmidrule{2-5}
\multirow{2}{*}{Loan, random} & oracle & 0.637 & 1.0 & $0.068_{\pm 0.081}$\\
& iCARMA & 0.963 & 0.6  & $0.08_{\pm 0.094}$ \\
\midrule
GSMC, all (4) & iCAMRA & 0.88 & 0.956 & $0.579_{\pm0.275}$ \\ 
GSMC, random & ICARMA & 0.753 & 0.886 & $0.549_{\pm0.256}$ \\ 
\bottomrule
\end{tabular}
}
\end{wraptable}

\subsection {How do hard individual constraints impact the key recourse desiderata?}
 We first investigate whether the quality of recourse solutions, as measured by the core metrics, degrades when users can \textit{only act on a strict subset of globally actionable features}. This situation can naturally arise if some globally actionable features cannot be acted on simultaneously, or if acting on all such features at once is overly demanding~\cite{Venkatasubramanian_2020}. Table~\ref{tab:exp_1_uniform} reports how the core metrics change as the number of individually actionable features decreases.

 \textit{Preference Configuration.} In the reference case, all globally actionable features are individually actionable (rows 1 and 6 of Table~\ref{tab:exp_1_uniform}). We compare this to fully uniform preference sampling, as applied during iCARMA training (rows 5 and 7), and to scenarios with a fixed number of individually actionable features (from 1 to all 4). In the fixed-number scenario, the specific features are sampled uniformly at random for each user, while the total number of actionable features remains constant across the deployment-time population.

\noindent  \textbf{In the absence of individual constraints, recourse is generally attainable.} When all four globally actionable features are also individually actionable for every user, amortized iCARMA closely matches the non-amortized Oracle both in terms of cost and (perfect) validity. As expected, Oracle achieves perfect plausibility but slightly declined validity because it enforces plausibility as a hard constraint, while iCARMA attains slightly lower plausibility because it uses plausibility as a regularizer. Notably, iCARMA provides recourse recommendations at scale much faster than Oracle. For example, to provide solutions over the whole population with three to four actionable features, Oracle requires over two hours, whereas a trained iCARMA model requires around ten minutes.

\begin{wrapfigure}[17]{r}{0.45\textwidth}
  \vspace{-12pt}
  \centering
  \includegraphics[width=0.45\textwidth]{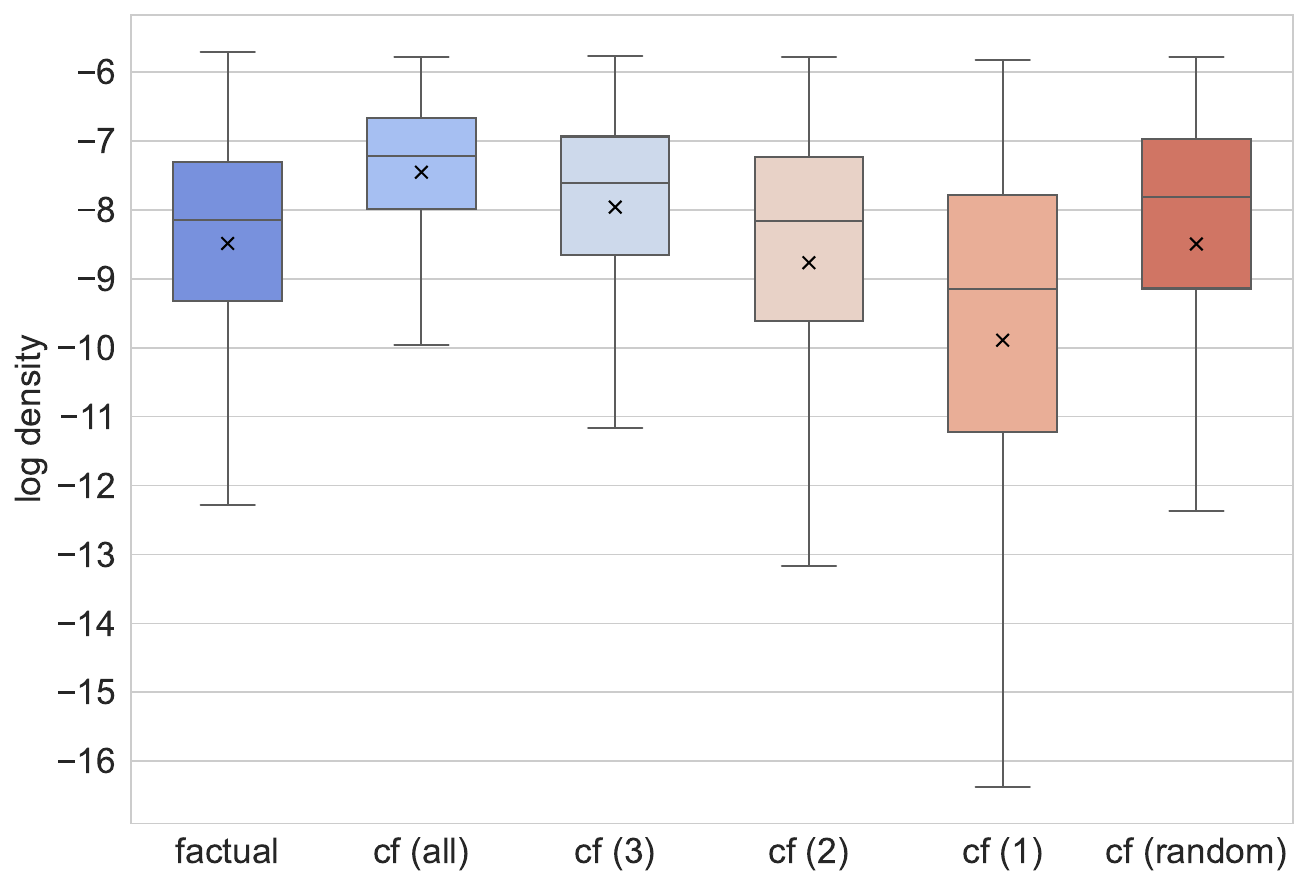}
  \caption{Distribution of the log densities for the factual and counterfactual (cf) under different numbers of actionable features (indicated in brackets).}
  \label{fig:exp_1_log_densities}
  \vspace{5pt}
\end{wrapfigure}

\noindent  \textbf{Validity-plausibility trade-off with fewer individual actionable features.}
As users impose stricter constraints by reducing the number of actionable features, \textit{both validity and plausibility degrade}, and this effect becomes \textit{more pronounced as fewer features remain available}. This trend holds for both iCARMA and the Oracle, indicating that even non-amortized solutions are affected. For the Oracle, validity drops sharply because it must always satisfy plausibility. For example, when only two features are actionable, the Oracle’s validity falls to nearly half of the base case with all four features. In Appendix~\ref{apx:add_res_2}, we ablate the impact of plausibility in the Oracle’s solutions and show higher validities but similar trends when the plausibility constraint is removed. In contrast, iCARMA exhibits a milder trade-off, preserving higher validity at the cost of less plausible counterfactuals. For instance, with only two actionable features, iCARMA’s plausibility is roughly half of the all-features-actionable case.

To further examine the interaction between hard constraints and distributional plausibility, Figure~\ref{fig:exp_1_log_densities} plots the log-density of recourse counterfactuals against that of the factual instances for different numbers of actionable features. A clear pattern emerges: if users have stricter constraints, meaning fewer individually actionable features, the log-density of the resulting counterfactuals systematically decreases, indicating lower plausibility. Moreover, the distribution of log-densities exhibits wider spreads toward low-density regions with reduced actionability.

\noindent  \textbf{Hard constraints have minor effect on cost.} On the other hand, the cost of valid recommendations, our primary optimization objective, remains largely comparable across constraint levels for both methods. The mean cost shows only a minor  to moderate increase as the number of individually actionable features decreases. 

\noindent  \textbf{Similar effects under causal-order-biased preferences.}
In Appendix~\ref{apx:add_res_1}, we additionally analyze non-uniform sampling of individually actionable features. Specifically, we assume that preferences align with the causal ordering of features. For example, individuals may prefer features that appear earlier or later in the causal graph, depending on their depth.  Such scenarios are particularly relevant for causal recourse, as feature depth can influence the resulting solutions~\cite{CARMA}. Across these settings, we observe similar trade-off patterns.

\noindent  \textbf{Takeaways.} These findings reveal a critical tension in AR that prior work has largely overlooked. Hard individual actionability constraints can render the recourse optimization problem over-constrained in many cases, substantially reducing both validity and plausibility.  Since individually actionable recommendations are not always attainable, future work should explore more nuanced socio-technical approaches. One promising direction is \textit{integrating recourse with social interventions}~\cite{von2022fairness}, which could distribute the burden between decision subjects and third parties such as governments or social planners. Such an approach would extend existing methods that improve fairness across demographic groups toward providing equal opportunity at the individual level. However, such interventions may also create incentives for users to exaggerate their constraints, highlighting the need for careful design and, as in strategic classification \cite{strategicClassification}, game-theoretic analysis.

\begin{table}[t]
\centering
\caption{Impact of \textbf{preference consideration as rankings} on recourse metrics across four cost profiles: concave, linear, convex, and constant (no cost individualization) for amortized iCARMA and non-amortized causal Oracle.}
\label{tab:exp_2_rank}
\resizebox{0.8\columnwidth}{!}{%
\begin{tabular}{lccccccc}
\toprule
\textbf{dataset+constraints} & \textbf{individual cost profile} & \textbf{method} & \textbf{validity} & \textbf{plausibility} & \textbf{cost}  & \textbf{hap} \\
\midrule
\multirow{8}{*}{Loan, soft } & \multirow{2}{*}{constant} & oracle & 0.959 & 1.0 &$0.057_{\pm 0.064}$ & 0.614 \\
& & iCARMA & 0.999 & 0.886 & $0.069_{\pm 0.077}$ & 0.852 \\
\cmidrule{3-7}
& \multirow{2}{*}{concave} & oracle & 0.959 & 1.0 & $0.074_{\pm 0.088}$ & 0.492 \\
& & iCARMA & 0.998 & 0.631 &$0.079_{\pm 0.088}$ & 0.044 \\
\cmidrule{3-7}
 & \multirow{2}{*}{linear} & oracle & 0.959 & 1.0 & $0.072_{\pm 0.085}$ & 0.443 \\
 & & iCARMA & 0.996 & 0.532 & $0.077_{\pm 0.086}$ & 0.029 \\
\cmidrule{3-7}
 & \multirow{2}{*}{convex} & oracle & 0.959 & 1.0 &$0.07_{\pm 0.081}$ & 0.387 \\
 & & iCARMA & 0.998 & 0.541 & $0.077_{\pm 0.084}$ & 0.044 \\
\midrule
\multirow{8}{*}{\begin{tabular}{@{}l@{}} Loan, hard \\ (random) \\ + soft \end{tabular}} & \multirow{2}{*}{constant} & oracle & 0.637 & 1.0 & $0.068_{\pm 0.081}$ & 0.694 \\
& & iCARMA & 0.963 & 0.6 & $0.08_{\pm 0.094}$ & 0.859 \\
\cmidrule{3-7}
 & \multirow{2}{*}{concave} & oracle & 0.637 & 1.0 & $0.079_{\pm 0.095}$ & 0.607 \\ 
 & & iCARMA & 0.951 & 0.506 & $0.081_{\pm 0.095}$ & 0.447 \\
 \cmidrule{3-7}
 & \multirow{2}{*}{linear} & oracle & 0.637 & 1.0 & $0.076_{\pm 0.091}$ & 0.596 \\ 
 & & iCARMA & 0.958 & 0.439 & $0.083_{\pm 0.098}$ & 0.440 \\
 \cmidrule{3-7}
 & \multirow{2}{*}{convex}  & oracle & 0.637 & 1.0 & $0.074_{\pm 0.088}$  & 0.583 \\
 & & iCARMA & 0.962 & 0.438 & $0.085_{\pm 0.103}$ & 0.436 \\
 \midrule
  GSMC, soft & linear & iCARMA & 0.865 & 0.879 & $0.618_{\pm0.305}$ & 0.30 \\ 
\midrule
\multirow{2}{*}{\begin{tabular}{@{}l@{}} GMSC, hard (3) \\ + soft \end{tabular}}  & constant & iCARMA & 0.76 & 0.88 & $0.55_{\pm0.25}$ & 0.40 \\ 
 & linear & iCARMA & 0.73 & 0.86 & $0.57_{\pm0.27}$ & 0.34 \\ 
\bottomrule
\end{tabular}
}
\end{table}

\subsection{ Do soft individual constraints lead to milder trade-offs in recourse optimization?}

Motivated by the severe trade-offs observed in our prior evaluation, we investigate how outcomes change when users express preferences as rankings or scores (see Section \ref{sec:prompting}) over globally actionable features. This setting captures \emph{soft constraints}, reflecting situations where individuals may prefer certain actions but do not completely rule out others. We consider scenarios in which users provide either rankings or scores as soft constraints. In addition, we examine cases where some features are rejected as individually non-actionable, thereby analyzing settings where both hard and soft constraints are present. Results for rankings are reported in Table \ref{tab:exp_2_rank}, and results for scores in Table~\ref{tab:exp_2_score}.

\noindent
 \textit{Preference Configuration.}
For purely soft constraints, we sample rankings uniformly at random over globally actionable features. When combining hard and soft constraints on the Loan dataset, we first sample the set of individually actionable features—both the number and their identities—uniformly, and then draw a uniform ranking over the selected features. For GMSC, we randomly exclude one feature as individually non-actionable and sample uniform rankings over the remaining features. We also consider purely soft, score-based constraints on the Loan dataset (Table \ref{tab:exp_2_score}), where each globally actionable feature is assigned a score sampled uniformly between 1 and 4. In this score-based setting, hard and soft constraints are combined by introducing an additional score to represent individual non-actionability. Finally, to evaluate the impact of different cost profiles, we compare the three profiles from Section \ref{sec:indiv_cost} (concave, linear, and convex) on the Loan dataset, while the constant profile reflects no influence of soft constraints during optimization.

\noindent
 \textbf{Ranking-based soft constraints reduce trade-offs but may still suggest hard actions.} From Table \ref{tab:exp_2_rank}, compared to the previous \emph{hard-constrained} case, we see that trade-offs in core recourse metrics \textit{are milder} for both iCARMA and Oracle, resulting in higher plausibility and validity. Notably, with individualized (non-constant) cost profiles, iCARMA recommends \emph{harder features} at substantially lower rates (lower hap values) than with constant cost profiles. For the non-amortized Oracle, the reduction in hap values is less pronounced, likely due to its strict enforcement of plausibility. These patterns hold across all cost profiles,  highlighting that individual-cost aware optimization can \textit{reduce the probability of recommending actions on the least preferred features}.

The non-zero hap values indicate that soft constraints alone cannot fully capture individual non-actionability. Consequently, if users only provide soft preferences, they risk receiving ineffective recommendations when their \textit{least-preferred features are actually non-actionable} rather than merely difficult to change. This underscores the importance of accurately understanding a user’s true preferences and actionable conditions.

\noindent
 \textbf{Additional hard constraints worsen trade-offs.}
For the \emph{hard + soft} setup in Table~\ref{tab:exp_2_rank}, compared to the purely soft case, validity drops remain mild for iCARMA, while hap values increase, indicating that recourse recommendations more often target harder features within the allowable set. Hence, the trade-offs reflect a middle ground in this individual actionability scenario.  However, this also indicates that auditing based solely on soft constraints can significantly overestimate the performance of a recourse recommendation system if individuals \textit{truly have hard constraints}.

\vspace{5pt}
\begin{wraptable}[13]{r}{0.5\columnwidth}
\centering
\vspace{-13pt}
\caption{Impact of \textbf{preference consideration as scores} on recourse metrics across cost profiles for iCARMA  {on the Loan Dataset}.}
\label{tab:exp_2_score}
\resizebox{0.5\textwidth}{!}{%
\begin{tabular}{p{1.5 cm}ccccccc}
\toprule
\makecell[l]{\textbf{constraint}} & \makecell{\textbf{individual}\\ \textbf{cost profile}} & \textbf{validity} & \textbf{plausibility} & \textbf{cost} & \textbf{hap} \\
\midrule
\multirow{4}{=}{soft}  & constant &  0.996 & 0.841 &$0.067_{\pm 0.07}$ & 0.829 \\
 & concave & 0.995 & 0.713 &$0.073_{\pm 0.078}$ & 0.225 \\
& linear &  0.991 & 0.646 & $0.074_{\pm 0.086}$ & 0.178 \\
& convex &  0.982 & 0.5 & $0.081_{\pm 0.082}$ & 0.194 \\
\midrule
\multirow{4}{=}{hard (random) + soft }  & constant & 0.991 & 0.712 & $0.073_{\pm 0.083}$ & 0.838 \\
& concave & 0.99 & 0.618 & $0.076_{\pm 0.087}$ & 0.415 \\
 & linear & 0.987 & 0.569 & $0.078_{\pm 0.097}$ & 0.349 \\
 & convex &  0.964 & 0.443 & $0.086_{\pm 0.092}$ & 0.323 \\ 
\bottomrule
\end{tabular}%
}
\end{wraptable}

\noindent
 \textbf{Cost profiles are more influential in score-based elicitation.}
 Table~\ref{tab:exp_2_score} shows the iCARMA results for score-based preference elicitation. The overall trends mirror the ranking case: moving away from purely hard individual actionability constraints moderates trade-offs with other metrics. However, cost profiles have a more pronounced effect for this setup. For example, a concave cost profile (where cost increases more mildly for lower-preference features) improves plausibility but also raises hap, whereas a convex profile (where cost rises steeply) reduces both plausibility and hap, reflecting fewer interventions on harder features but at the expense of less distributionally plausible counterfactuals.

\noindent
\textbf{Takeaway.} We observe the need for sufficiently expressive user prompting that can identify both hard and soft constraints and clearly distinguish between them. Considering only soft constraints may conceal true hard constraints, leading to an overestimation of recourse performance. Conversely, users must understand the prompting scheme to avoid unnecessarily overconstraining the optimization problem, which could impede the effectiveness of recourse. Future work should also explore combining our approach with additional post-hoc prompting to further enhance expressivity and to study the impact of other potential constraints.

\subsection{Case Study: Can accounting for individual actionability reveal underlying recourse unfairness?}

Finally, using the semi-synthetic Loan dataset, we design a case study to examine how heterogeneous individual actionability constraints can affect the fairness of algorithmic recourse. We hypothesize that \textit{systematic differences} in these constraints across demographic groups can create disparities beyond those stemming from causal downstream effects of sensitive group membership on features~\cite{von2022fairness}, causing some groups to receive recommendations that are costlier or less plausible. Importantly, these disparities are \textit{not caused by accounting for individual actionability} but \textit{revealed by it}. Therefore, ignoring individual actionability may often leave certain groups with ineffective recourse that existing fairness studies would fail to capture.

\begin{figure}[t]
  \centering
\includegraphics[width=\textwidth]{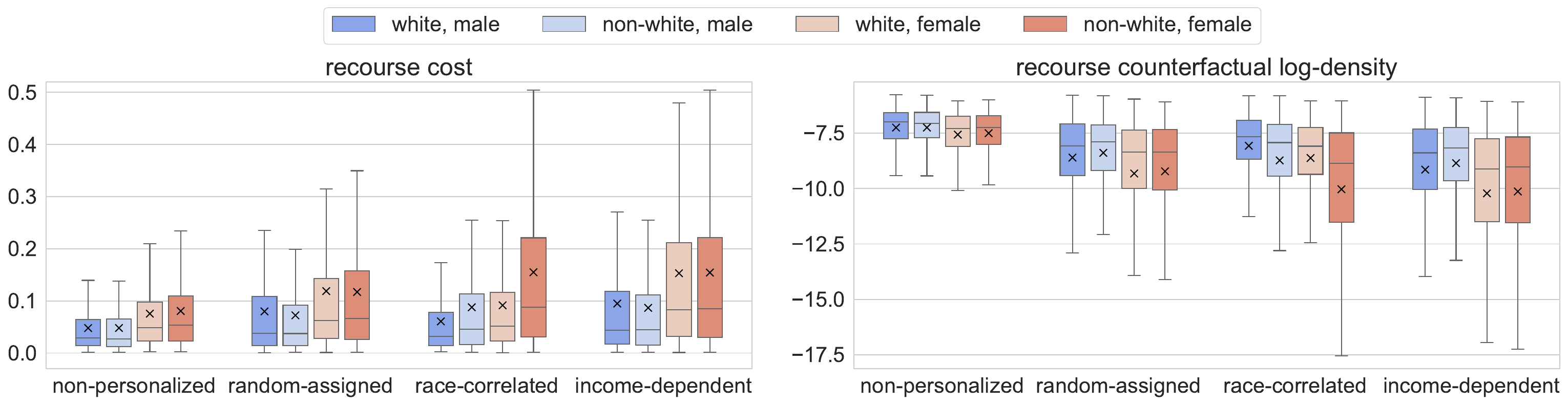}
\caption{Cost (left) and distributional log-density (right) results of recourse recommendations across intersectional gender and race groups for four scenarios: \textbf{(1) non-personalized:} all globally actionable features are equally actionable, \textbf{(2) random-assigned:} privileged preference profiles assigned randomly, \textbf{(3) race-correlated:} 90\% of white, 5\% of non-white individuals assigned the privileged profile, \textbf{(4) income-dependent:} top-earning 25\% of individuals assigned the privileged profile.}
  \label{fig:exp_3_fairness}
\end{figure}

\noindent \textit{Preference Configuration.}
We consider two demographic axes for our fairness analysis. First, we use the existing \emph{gender} attribute to examine how heterogeneous actionability constraints can reinforce pre-existing recourse disparities arising from downstream causal effects on other features. Second, we introduce a stylized \emph{race} attribute, randomly assigned to individuals. In this setup, race does not affect other features but only (potentially) influences an individual’s actionability profile. We ensure that the dataset maintains equal proportions across all gender-race combinations.
We define two stylized preference profiles: \emph{privileged} and \emph{non-privileged}. We consider \emph{hard actionability constraint scenario} where all individuals are restricted to modifying only two features, but the groups differ in which features are likely to be actionable. The privileged group is biased toward having harder-to-change financial features (Income, Savings), while the non-privileged group is biased toward easier-to-change features (Loan Amount, Duration). To keep the analysis focused, we do not consider any specific cost profile. \footnote{These stylized setups may not reflect real-world scenarios, but are intended to illustrate how individualized actionability can reveal recourse disparities.}
We compare these scenarios to a baseline in which all globally actionable features are also individually actionable. Preference profiles are assigned in three ways: \emph{randomly}, \emph{income-dependent} (using the top 25\% income as a threshold), or \emph{race-correlated} (90\% of white individuals and 5\% of non-white individuals assigned privileged). 
\\
\\We train iCARMA using uniform random preferences and, for each scenario, evaluate recourse using unweighted $\ell_2$ cost and counterfactual log density to assess impacts on cost and plausibility across gender and race. Figure~\ref{fig:exp_3_fairness} shows the results.

\noindent \textbf{Heterogeneous preference profiles can amplify existing disparities.}
Comparing the \textit{randomly assigned} preferences to the baseline (no-preference) scenario across gender, we observe that even randomly assigned individualized actionability constraints can amplify existing recourse disparities. As we see in Figure~\ref{fig:exp_3_fairness}, for the baseline, female users already face higher recourse costs and less plausible recommendations (lower log-densities) compared to male users. These gaps increase when individual actionability constraints are considered.
The effects are more pronounced under \emph{income-dependent preferences}. Gender and income are correlated, so incorporating individualized actionability constraints makes recourse for female applicants substantially more costly and less plausible. Hence, accounting for preferences can not only expose existing disparities, but it can also reveal that they are larger than previously recognized.

\noindent \textbf{Individualized actionability can uncover hidden disparities.}
Our results also highlight how ignoring individual actionability can mask potential disparities. Under \textbf{race-correlated actionability constraints}, we observe cost and plausibility gaps emerging solely from individualized actionability constraint profiles. Even though race does not influence other features or the label, if it shapes which features are actionable, it can lead to higher costs and less plausible recommendations for certain groups.

\noindent \textbf{Individualized actionability can introduce intersectional disparities.}
Accounting for individual actionability can reveal disparities resulting from interaction effects across multiple demographic axes. As shown in Figure~\ref{fig:exp_3_fairness}, these effects are strongest when individual constraints correlate with demographic factors. In our case study, even when preferences are only race-correlated, they interact with existing gender disparities, disproportionately affecting non-white females with higher-cost and less plausible recourse recommendations.

\noindent \textbf{Takeaway.} 
Our results show that ignoring individual actionability constraints can lead to substantial underestimation of existing disparities and to overlooking unmeasured sources of inequity in AR solvers. 
This could ultimately cause recourse to backfire. While AR aims to enhance user agency and improve the acceptability of algorithmic systems, ignoring individual constraints may leave users worse off, bearing blame for not overturning a negative decision without having been given individually implementable options. This risks reframing structural disparities as individual failures, further marginalizing disadvantaged groups and eroding trust in ADS over time. Our findings caution against purely technical solutions and highlight the need for interdisciplinary and socio-technical approaches.

\section{Limitations, Broader Implications, and Conclusion}

Our work formalizes individual actionability constraints and demonstrates through empirical studies that effective recourse systems must account for heterogeneity in user preferences and capabilities. At the same time, several challenges remain for both research and practice.

Deploying causal recourse in real-world settings requires \textbf{robustness to incomplete causal knowledge, hidden confounders, and reliable evaluation data}. Users’ preferences are typically unknown a priori and must be elicited carefully, raising design challenges that call for interdisciplinary collaboration across psychology, human–computer interaction, and decision sciences.
Similarly, \textbf{metric and objective design is critical for recourse.} Standard measures like $\ell_2$ may not correctly capture perceived effort~\cite{heidari2019long}---for instance, they cannot account for recourse recommendations targeting individually non-actionable features being effectively useless. Moreover, misaligned optimization objectives can unintentionally induce disparities~\cite{perello2025discrimination}. These issues highlight the importance of user-centered metric definitions and studying individual actionability tradeoffs through such measures to enable practical deployment.

 Our case studies demonstrate that \textbf{algorithmic recourse alone is insufficient to ensure equitable outcomes}. Trade-offs between feasibility, plausibility, and effort may shift the burden of structural limitations onto individuals in inequitable ways. Integrating recourse with \textbf{broader social interventions}—such as shared responsibility frameworks involving governments or social planners \cite{von2022fairness}—could enable universally attainable individualized recourse and distribute responsibility more fairly. %
However, social interventions involve reallocation of public resources and thus require \textbf{public and political engagement}.
In addition, such approaches face the challenge of ensuring faithful representation of individual constraints by mitigating potential strategic behavior, thereby enabling a fair distribution of burden among the involved parties.

Finally, several important avenues remain for advancing personalized actionable recourse. First,  more expressive prompting schemes, post-hoc feedback, or conversational interfaces may improve the elicitation of individual actionability constraints. Moreover, it is essential to consider potential \textit{trade-offs} with recourse objectives and other trustworthiness aspects, such as privacy~\cite{pawelczyk2022privacyrisksalgorithmicrecourse} and robustness~\cite{pawelczykprobabilistically,Lakkaraju_21_1,dominguez2022adversarial}. More broadly, holistic studies of personalization are needed to ensure that recourse systems—and automated decision-making systems more generally—are actionable, fair, and socially responsible.

\section*{Generative AI Usage Statement}
The core ideas, proposed methodologies, and experiments were developed solely by the authors. AI tools such as ChatGPT and Grammarly were used only as assistive tools in the preparation of the manuscript. Specifically, these tools were used to help revise parts of the text for grammatical and structural correctness. However, these tools were not used to generate any text directly.

\section*{Acknowledgments}
This work has been supported by the project “Society-Aware
Machine Learning: The Paradigm Shift Demanded by Society to Trust Machine Learning,” funded
by the European Union and led by IV (\texttt{ERC-2021-STG, SAML, 101040177}); and the Deutsche
Forschungsgemeinschaft (DFG) grant number \texttt{389792660} as part of the Transregional Collaborative
Research Centre TRR 248: Center for Perspicuous Computing (CPEC) (\texttt{TRR 248 – CPEC}).  Views and opinions expressed are, however, those of the author(s) only and do not necessarily reflect
those of the aforementioned funding agencies. Neither of the aforementioned parties can be held
responsible for them.

\bibliographystyle{ACM-Reference-Format} 
\bibliography{references}  

\appendix
\section{Background}

\subsection{Extended Related Work}
\label{RelWork_Appendix}

Wachter et al.~\cite{wachter2017counterfactual} introduced \textbf{counterfactual explanations}, combining ideas from adversarial perturbations and philosophy of science. A counterfactual explanation identifies the minimal changes needed to obtain a different outcome, for example, indicating that a slightly higher income would have led to loan approval. In this context, Wachter et al. already acknowledged that \textit{``knowing the smallest possible change to a variable or set of variables to arrive at a different outcome may not always be the most helpful type of counterfactual. Rather, relevance will depend also upon other case-specific factors, such as the mutability of a variable or real-world probability of a change.''} \cite{wachter2017counterfactual}. 
Subsequent research has extended the counterfactual framework toward recourse recommendations~\cite{karimi2022survey, valera2020survey}, supplementing the native optimization problem with additional constraints or objectives accounting for further desiderata such as actionability~\cite{10.1145/3287560.3287566}, feasibility~\cite{10.1145/3287560.3287569, pmlr-v108-karimi20a, 10.1145/3375627.3375850} or plausibility~\cite{kanamori2020dace,joshi2019realisticindividualrecourseactionable,Pawelczyk_2020}. 

Recourse recommendations aim to identify changes that are both realistic and practically attainable. This distinguishes them from purely explanatory counterfactuals~\cite{karimi2022survey}, which may involve immutable attributes such as race or gender~\cite{wachter2017counterfactual}. Ensuring genuine actionability requires a shift from nearest counterfactual explanations to \emph{minimal interventions}, indicating that recourse consists of the set of least-effort, causally effective actions rather than residing in the space of counterfactuals~\cite{karimi2021algorithmic}.
Recent advancements have focused on extending validity guarantees over time~\cite{Temporal_AR_Ceccon, Temporal_AR_deToni, Temporal_AR_Fonseca, The_Game_Of_Recourse}, making the generation of recourse recommendations more efficient~\cite{mahajan2020preservingcausalconstraintscounterfactual, pmlr-v180-nemirovsky22a, verma2022amortized, de2023synthesizing, CARMA}, and systemtically tailoring recourse to individual users' needs \cite{2022personalized_pref_eliction, yadav2021low,Yetukuri_2023}.

\textit{Personalization of Recourse.}
As personalizing recourse directly refers to incorporating an individual's constraints and needs, it is essential to at least include psychological insights on what satisfies users with respect to recourse \cite{Keane_2021}. 
Studies indicate that while counterfactual-style explanations are generally perceived as more actionable than feature-importance explanations (especially after unfavorable outcomes) \cite{Langer_2024, Singh_2023, Shulner_Tal_2022}, \textit{counterfactuals misaligned with user needs may be worse than feature-based explanations} \cite{upadhyay2025counterfactual}. Further studies also suggest that users prefer personalized recourse plans over generic ones \cite{Wang_2023}, and that the opportunity to express preferences increases user satisfaction \cite{Uth_2025}. Yet, the interpretation of these findings requires caution, as the mere opportunity to “have a voice” can already improve fairness perceptions~\cite{Hellwig_2023}.

Studies suggest that \emph{actionability itself} plays a central role, even if adherence to users' expressed preferences does not consistently yield more favorable reactions \cite{Uth_2025}. \citet{koh24} find that providing action constraints---such as prohibiting changes to certain features or specifying actionable ranges---benefits participants more than assigning feature priorities. Action constraints reduced frustration, whereas priority constraints increased perceived mental demand and effort. Similarly, \citet{tomu25} show that the perceived reasonableness of a recourse recommendation strongly correlates with final acceptance of an automated decision, while perceived actionability shows a positive association only when it is sufficiently low, indicating that users are especially sensitive to whether they can execute the recommended actions at all.

\subsection{Structural Causal Models}
\label{SCM_Appendix}

Structural Causal Models (SCMs) as introduced by J. Pearl \cite{pearl2009causal1, pearl2009causality2} offer a practical framework to model dependencies between features that a recourse generating system is required to capture and thus serve as the standard operationalization of causality within the field \cite{CARMA, karimi2021algorithmic}.
An SCM consists of observed (endogenous) variables $X$, unobserved (exogenous) variables $U$, and a set of functions\footnote{We restrict the more general case of structural equations in line with~\cite{CARMA} to functions.} $F$.
\\The \textbf{observed (endogenous) variables $X$} constitute the input features provided to the automated decision-making system. In alignment with the assumptions by~\cite{CARMA}, we assume that causal relationships between those variables can be modeled by a directed acyclic graph (\textit{causal graph}). If there is an outgoing edge from $x_1$ to $x_2$ in the causal graph, i.e., if $x_1$ is considered to 'causally influence' $x_2$, $x_1$ is called a \textit{causal parent} of $x_2$.
\\ One way to get an intuition about \textbf{unobserved (exogenous) variables $U$} is to regard them as encoding variance not otherwise explained by the model. Typically, the value of endogenous variables can not be fully attributed to their causal parents. To account for the remaining unexplained variance in an endogenous variable $x_i$, there is an exogenous variable $u_i$ capturing that unexplained variance. We require the $u_i$ to be mutually independent - a condition called \textit{Causal Sufficiency.}\footnote{This especially indicates that there are no \textit{hidden confounders}. One can think of a hidden confounder as a causal parent ('common cause') of multiple exogenous variables. A classic example is sunny weather as a common cause of sunburn and ice cream sales, rendering ice cream consumption to be positively correlated with sunburn rates, even so obviously not causing them to rise.}
The \textbf{set of functions $F$} allows obtaining $X$ given $U$ (\textit{abduction}), while the set of inverted functions $F^{-1}$ allows obtaining $X$ given $U$ (\textit{prediction}). We assume the value of each $x_i$ to be determined by a differentiable\footnote{The structural functions as well as their inverses are both required to be differentiable.} function $f_i$ of $u_i$ and the causal parents of $x_i$. 

\subsection{Causality-aware Generation of Counterfactuals based on SCMs}
Given an SCM $M$, counterfactuals can be generated by a three-step procedure: \textit{abduction - intervention - prediction} \cite{pearl2009causal1, pearl2009causality2, CausalNF}.
One way to get an intuition about interventions is to see them as simulating an idealized experiment, where only a single factor contributing to an outcome of interest is altered to observe how that change propagates along 'causal pathways'. We define an \textbf{intervention} on $x_i$ as applying the \textit{do-operator} $do(x_i = a_i)$ which modifies the SCM. The modified SCM $M'$ is identical to $M$ except that $x_i$ is no longer determined by the structural function $f_i$ but is constant and determined by the interventional value $a_i = u_i = x_i$. 
Thus, a recourse recommendation, i.e., a set of actions, is formalized as a set of interventions $ A = \{ \{x_i := a_i  \}_{i\in I} \}$, 
with $I$ being the set of features selected for intervention. The corresponding (structural) counterfactual $x^{CF} := x^{SCF}(A,x)$ can be determined by prediction based on the SCM modified according to the interventions in $A$. 
Given the exogenous variables, a recourse model should compute a minimal (or near-minimal) set of interventions, such that the predicted counterfactual results in a positive outcome of the decision-making classifier.

\subsection{Causal Normalizing Flow}
\label{CNF}
Under the assumption of causal sufficiency, an SCM's structural functions can be modeled as autoregressive Triangular Monotonic Increasing (TMI) maps, allowing Causal Normalizing Flow to build on \textit{Autoregressive Normalizing Flows} \cite{ANF_1, ANF_2}, which are able to approximate any TMI map arbitrarily well~\cite{CausalNF}. 
\\Given the causal graph of a Structural Causal Model (SCM) $M$, and assuming sufficient training data originating from the causal process modeled by $M$, the latent space learned by CNF corresponds to the exogenous space of $M$ (\textit{causal consistency}). Thus, the transformation from input space to latent space enables abduction, while the inverse transformation from latent space back to input space allows for prediction. In addition, CNF implements the do-operator to perform interventions. As CNF is not explicitly operating on the structural functions but, like a ('non-causal') Normalizing Flow~\cite{NF}, on probability distributions, the analog to the modified SCM given an action A is the \textit{interventional exogenous distribution}, defined by
\begin{equation}
    Z^A = \prod_{i \in I} \delta (\{ X_i := a_i\}) \cdot \prod_{j \notin I} p_j(z_j) 
\end{equation}
As the features that have been intervened on are set to a fixed value, their probability distribution is determined by a Dirac $\delta$. All other variables in the causal latent space\footnote{To distinguish the true exogenous from the causal latent space, we use $u_i$ to denote the exogenous variable, while $z_i$ denotes the variable in the causal latent space.}, denoted by $z_j$, keep the learned distribution. 
Assuming the latent variables are mutually independent\footnote{If we had not required causal sufficiency, the exogenous variables might not behave accordingly, causing the exogenous space to be distinct from CNF's latent space.}, the joint distribution over the latent space factorizes into the product of the marginal distributions of the individual latent variables.

\subsection{CARMA Optimization Pipeline}
\label{carma}

As the \textit{"\textbf{C}ausal \textbf{A}lgorithmic \textbf{R}ecourse framework utilizing neural network \textbf{M}odel-based \textbf{A}mortization"} (\textbf{CARMA})~\cite{CARMA} is trained in an unsupervised setting, it does not require access to ground-truth (optimal) recourse actions during training. Given a set of negatively classified samples and white-box access to a decision-making classifier $h(x_u)$ as well as a pre-trained Causal Normalizing Flow (to incorporate causal reasoning), the Mask Network (with parameters $\omega$) and the Action Network (with parameters $\phi$) of CARMA are jointly trained by optimizing a \textbf{combined loss} $L_{\omega,\phi}(x_u, x_u^{CF})$ defined by 
\begin{equation}
\label{CARMA_native_loss}
   L_{\omega,\phi}(x_u, x_u^{CF}) = \min_{\omega, \phi}[L^{mask}_{\omega}(x_u)+ L^{action}_{\phi}(x_u, x_u^{CF})] + \lambda \cdot HingeLoss (h(x_u^{CF}); 1, \beta)    
\end{equation}
\noindent where the hinge loss term constrains for validity, ensuring a positive decision of the decision-making classifier on the counterfactual. The hyperparameter $\beta$ determines the margin distance, with larger values encouraging the classifier's positive prediction probability to be larger. The \textbf{Action Loss} $L^{action}_{\phi}
(x_u,x_u^{CF})$, defined by 
\begin{equation}
L^{action}_{\phi}
(x_u,x_u^{CF}) = \min_{\phi} \sum^{k}_{i =1} \mathbb{I}_{i\in I}\cdot (x_{u,i} -x^{CF}_{u,i})^2
\label{Action_loss}
\end{equation}
\noindent optimizes for cost measured in terms of $\ell_2$-distance between the normalized factual and counterfactual for any globally actionable feature $i$ such that $\mathbb{I}_{i\in I} = 1$, i.e., the feature was selected by the Mask Network for intervention. This ensures that only active changes contribute to the cost, while downstream causal effects do not. 
The \textbf{Mask Loss} $L^{mask}_{\omega}(x_u)$ is defined by 
\begin{equation}
L^{mask}_{\omega}(x_u) = \min_{\omega} \sum^{k}_{i =1}D_{KL}\left( Bernoulli(\mu_i|z_u)||Bernoulli(\pi_{i})\right)  
\label{Mask_loss}
\end{equation}
\noindent where the factual in the causal latent space, $z_u$, is obtained by the pre-trained Causal Normalizing Flow from $x_u$ and $Bernoulli(\mu_i|z_u)$ denotes the probability distribution learned by the Mask Network to intervene on feature $i$ given $z_u$. 
By default, the Mask Loss serves primarily as an entropy regularization, as $\pi_i=0.5$ for all $i$.

\subsection{Population-level Bias in CARMA}
As shown in~\cite{CARMA}, CARMA allows to set distinct per-feature probabilities $p_i$ on population level, introducing a bias towards intervening on particular features more likely than on others. Also, the implementation allows the Action Network to minimize a population-level weighted $\ell_2$-distance, rather than the standard $\ell_2$-distance-based cost. In this biased mode, the Action Loss $L^{bias-action}_{\phi}(x_u, x_u^{CF})$ becomes defined by
\begin{equation}
L^{bias-action}_{\phi}(x_u, x_u^{CF}) = \min_{\phi} \sum^{k}_{i =1} \mathbb{I}_{i\in I} \cdot w_i \cdot (x_{u,i} -x^{CF}_{u,i})^2
\label{weighted_Action_loss}
\end{equation}
where $w_i$ denote \textbf{population-level weights} capturing that some features are, \textit{for all users}, more difficult to change than others.

\section{iCARMA: Implementation Details}
\label{iCARMA}

\subsection{Modified Loss Functions}
\label{iCARMA_loss_modified}

To inform the optimization of individual users' needs, we introduce individual actionability weights $w_u$ to the Action Loss and actionability priors $\pi_u$ to the Mask Loss. 
The \textbf{individualized Action Loss} $L^{i-action}(x_u, x_u^{CF}, s_u)$ therefore becomes defined by 
\begin{equation}
\label{new_action_loss}
 L^{i-action}(x_u, x_u^{CF}, \text{\textcolor{red}{$s_{u}$}}) = \min_{\phi} \sum^{k}_{i =1} \mathbb{I}_{i\in \text{\textcolor{red}{$I_{u}$}}} \cdot \text{\textcolor{red}{$w_{u,i}$}} \cdot (x_{u,i} -x^{CF}_{u,i})^2
\end{equation}

\noindent where the \textbf{actionability weights} $w_{u,i}$ are systematically determined by the actionability score $s_{u,i}$. In line with the adapted cost function introduced in Section \ref{sec:indiv_cost}, we obtain $w_{u,i}$ from $s_{u,i}$ by
\begin{equation}
\label{w}
w_{u,i} = w^{max} - (w^{max}- w^{min})\cdot \left(1-\left(\frac{s_{u,i}-1}{s_u^{max} -1}\right)^{\alpha}\right)
\end{equation}

\noindent with 
\begin{align}
s_{u}^{max}= \begin{cases} 2 & \text{if }\max_{i \in AF_u}(s_{u,i})=1 \\\max_{i \in AF_u}(s_{u,i})& \text{else }\end{cases}
\end{align}

\noindent and given the hyperparameters $w^{max}$, $w^{min}$ and $\alpha$, which jointly determine how strong the penalty increases for increasing $s_{u,i}$.
\\By $I_u$, we denote the set of features selected by the Mask Network for intervention, given the r\textbf{estriction to features individually actionable} for $u$. Thus, while $I \subset AF$, we require $I_u \subset AF_u$, which might be more strict, as $AF_u \subset AF$.
\\
\\
We modify the Mask Loss accordingly by introducing user-level actionability priors $\pi_u$. Thus, the \textbf{individualized Mask Loss} $L^{i-mask}(x_u, s_u)$ becomes defined by 

\begin{equation}
\label{new_mask_loss}
L^{i-mask}(x_u, \text{\textcolor{red}{$s_{u}$}}) = \min_{\omega} \sum^{k}_{i =1}D_{KL}\left( Bernoulli(\mu_i|z_u)||Bernoulli(\text{\textcolor{red}{$\pi_{u,i}$}})\right)    
\end{equation}

\noindent where the \textbf{actionability priors} $\pi_{u,i}$ are systematically determined by the actionability score $s_{u,i}$.
To align the actionability priors with the actionability weights, we propose to obtain $\pi_{u,i}$ from $s_{u,i}$ by

\begin{equation}
\label{pr}
\pi_{u,i} = \pi^{min} + (\pi^{max} - \pi^{min})\cdot \left(1-\left(\frac{s_{u,i} -1}{s_u^{max}-1}\right)^{\alpha}\right)
\end{equation}

\noindent where $\pi^{max}$ and $\pi^{min}$ are hyperparameters determining the range of priors and, jointly with $\alpha$ how strong interventions on unpreferred features are penalized and $s_u^{max}$ is defined as above.
\\
\\Furthermore, we explicitly account for \textbf{plausibility} by adding a regularization term of the form
 \begin{equation}
 \label{Plausibility_regularization}
    \lambda_{P} \sum_{u=1}^{N} ReLU (log p(x_u)-logp(x^{CF}_u))
 \end{equation}

\noindent to the CARMA loss function, where $logp$ denotes the log-likelihood and $N$ the number of samples for which the loss is calculated, and $\lambda_{p}$ is a hyperparameter to be tuned.
We also explicitly account for \textbf{global feasibility} by adding a regularization term of the form
 \begin{equation}
 \label{Feaasibility_regularization}
    \lambda_{F} \sum_{u=1}^{N} \sum_{i=1}^{k_u} \left( \mathbb{I}_{x_{u,i}^{CF} > max_i} \cdot \frac{x_{u,i}^{CF} -max_i}{max_i} + \mathbb{I}_{x_{u,i}^{CF} < min_i} \cdot \frac{min_i - x_{u,i}^{CF}}{max_i} \right)
 \end{equation}

\noindent where $min_i$ denotes the minimum value of feature $i$ in a reference set and $max_i$ denotes the maximum value of feature $i$ in a reference set. The regularization thus introduces a linear penalty on the normalized exceeding of the globally feasible interval. We use the training set for reference, as we assume this to give a reasonable approximation of what is globally feasible. 

\subsection{Sampling Actionability Scores}
\label{Sampling}

Actionability scores can be sampled \textit{on the fly} in a data-augmentation manner each time a factual instance is loaded, or they can be pre-generated and supplied alongside the factual data, with a hybrid approach also being supported.
\textbf{General Score sampling} is implemented based on PyTorch Categorical distributions. Given a predefined categorial distribution over score values $[1,...,s^{max}]$, for each globally actionable feature $i$, $s_i$ is sampled independently from that categorial distribution. For each globally non-actionable feature j, $s_j = 0$.\footnote{This slightly deviates from the formalism introduced in Section \ref{sec:indiv_cost}, where both globally and individually non-actionable features were assigned a score of $s^{max}$. Here, the distinction between these two cases is introduced solely for implementation purposes.}
ICARMA supports defining up to four different distributions for each actionable feature, to allow for two different groups split along some (protected) attribute for fairness evaluation, each for train/validation and test separately. If not explicitly specified, no group distinction is made, and all probabilities are set uniformly.
\\We trivially obtain an implementation for \textbf{binary actionability scores} by either distributing the probability mass solely among the scores $1$ and $s^{max}$ or taking a rank sample and applying a constant cost profile during optimization (what was done in our implementation).
\\\textbf{Ranking-based actionability scores} are sampled in a two-step procedure.
First, a permutation of $[1, \dots, k]$ is sampled according to a predefined Categorial distribution\footnote{Precisely, we sample an index selecting the permutation from a list of all possible permutations.} to give a preliminary ranking among features. Then, to sample the cut-off for individual non-actionability, a threshold is sampled, and every entry in the preliminary ranking that is greater than or equal to the threshold is set to $s^{max}=k+1$. For the $l^{th}$ globally actionable feature $i$, $s_i$ is obtained by the $l^{th}$ entry of the thresholded preliminary ranking. For all globally non-actionable feature j, $s_j = 0$. 
Analogous to the score option, four different distributions (two groups for test and train each) can be customized for sampling the permutations and thresholds.

\section{Practical Details}
\label{Appendix_PracticalDetails}
This section outlines key practical details of our evaluation setup, detailing datasets, downstream classifiers, causal approximation, and recourse recommendation generation. Our implementation uses CPU only. We run all experiments on a MacBook Pro with M3 and 24 GB RAM.

\subsection{Preference Distributions and Sampling}
\label{Appendix_PrefDist_Sampling}

\begin{table}
\centering
\small
\caption{\textbf{Different distributions used for preference sampling on the Loan dataset. }Each row reports the probability of sampling a particular assignment of preference scores $pref_{u,i}$ to the globally actionable features LA, Dur, Inc, and Sav under a given configuration. The distribution biased according to causal order is denoted by co, while the distribution biased according to reverse causal order is denoted by rco. The distributions used to sample the privileged and non-privileged preference profiles for the fairness case study are reported in the last two columns.}
\label{tab:preference_probs}
\begin{tabular}{c c c c | c c c | c c}
\hline
$pref_{u,i} = 1$ & $pref_{u,i} = 2$  & $pref_{u,i} = 3$  & $pref_{u,i} = 4$  & uniform & co & rco &
\makecell{privileged\\pref profile} &
\makecell{non-privileged\\pref profile} \\
\hline
LA  & Dur & Inc & Sav & $1/24$ & $1/24$ & 0 & 0 & $1/6$ \\
LA  & Dur & Sav & Inc & $1/24$ & $1/24$ & 0 & 0 & $1/6$ \\
LA  & Inc & Dur & Sav & $1/24$ & $1/6$ & 0 & 0 & $1/24$ \\
LA  & Sav & Dur & Inc & $1/24$ & $1/24$ & 0 & 0 & $1/24$ \\
LA  & Inc & Sav & Dur & $1/24$ & $1/6$ & 0 & 0 & $1/24$ \\
LA  & Sav & Inc & Dur & $1/24$ & $1/24$ & 0 & 0 & $1/24$ \\
Dur & LA  & Inc & Sav & $1/24$ & 0 & $1/24$ & 0 & $1/6$ \\
Dur & LA  & Sav & Inc & $1/24$ & 0 & $1/24$ & 0 & $1/6$ \\
Inc & LA  & Dur & Sav & $1/24$ & $1/6$ & 0 & $1/24$ & 0 \\
Sav & LA  & Dur & Inc & $1/24$ & 0 & $1/24$ & $1/24$ & 0 \\
Inc & LA  & Sav & Dur & $1/24$ & $1/6$ & 0 & $1/24$ & 0 \\
Sav & LA  & Inc & Dur & $1/24$ & 0 & $1/24$ & $1/24$ & 0 \\
Dur & Inc & LA  & Sav & $1/24$ & 0 & $1/24$ & 0 & $1/24$ \\
Dur & Sav & LA  & Inc & $1/24$ & 0 & $1/6$ & 0 & $1/24$ \\
Inc & Dur & LA  & Sav & $1/24$ & $1/24$ & 0 & $1/24$ & 0 \\
Sav & Dur & LA  & Inc & $1/24$ & 0 & $1/6$ & $1/24$ & 0 \\
Inc & Sav & LA  & Dur & $1/24$ & $1/24$ & 0 & $1/6$ & 0 \\
Sav & Inc & LA  & Dur & $1/24$ & 0 & $1/24$ & $1/6$ & 0 \\
Dur & Inc & Sav & LA & $1/24$ & 0 & $1/24$ & 0 & $1/24$ \\
Dur & Sav & Inc & LA & $1/24$ & 0 & $1/6$ & 0 & $1/24$ \\
Inc & Dur & Sav & LA & $1/24$ & $1/24$ & 0 & $1/24$ & 0 \\
Sav & Dur & Inc & LA & $1/24$ & 0 & $1/6$ & $1/24$ & 0 \\
Inc & Sav & Dur & LA & $1/24$ & $1/24$ & 0 & $1/6$ & 0 \\
Sav & Inc & Dur & LA & $1/24$ & 0 & $1/24$ & $1/6$ & 0 \\
\hline
\end{tabular}
\end{table}

As we lack access to true user feedback, we sampled all preference scores ($pref_{u,i}$) for our experiments according to Section~\ref{Sampling}. During training, unless stated otherwise, preferences were sampled uniformly at random in a data-augmentation fashion. At test time, we used pre-sampled files to supply preferences. 

For the Loan dataset, the distributions used to parameterize the sampling procedure for rankings described in Section~\ref{Sampling} are given in Table~\ref{tab:preference_probs}. In addition to the uniform distribution, we considered a distribution biased according to causal order, i.e., towards having Income (Inc) and Loan Amount (LA) among the top two preferences (denoted by co), and a distribution biased according to reverse causal order, i.e., towards having Duration (Dur) and Savings (Sav) among the top two preferences (denoted by rco).
For the hard-constraint case, we used ranks sampled from one of the distributions given in Table~\ref{tab:preference_probs}, together with a constant cost profile and either uniform sampling of the individual actionability threshold or deterministic setting of the individual actionability threshold to a particular value between 2 (each user having only the highest-ranked globally actionable feature available) and 5 (each user having all globally actionable features available). For the soft and mixed constraint cases, we sampled ranks in the same way but made explicit use of the ranking by employing a non-constant cost profile.
For the score case, we used two different sampling configurations: either uniform sampling of scores from $[1,5]$ for the uniform case with individual non-actionability, or uniform sampling of scores from $[1,4]$ when no hard individual actionability constraints were assumed.
For the fairness case study, we sampled ranks from the distributions defining the privileged preference profile and the non-privileged preference profile, respectively, while the individual actionability threshold was not sampled but deterministically set to 3. As in the hard-constraint case, the cost profile was chosen to be constant. The privileged profile is biased toward having harder-to-change financial features (Income, Savings), while the non-privileged profile is biased toward easier-to-change features (Loan Amount, Duration). 

For the real-world experiments on the GiveMeSomeCredit dataset, we sampled ranks uniformly at random for all experiments. Depending on the configuration, the individual actionability threshold was either sampled uniformly at random from $[2,5]$, deterministically set to 5 (making all features available for all users), or set to 4 (making the top three features available for all users). Depending on the type of constraints, either a constant or linear cost profile was used.

\subsection{Semi-synthetic Loan Dataset}
\label{loan_Functions}
\subsubsection{Dataset Details}

\begin{figure}[h]
\vspace{-10pt}
  \centering
  \includegraphics[width=0.3\textwidth]{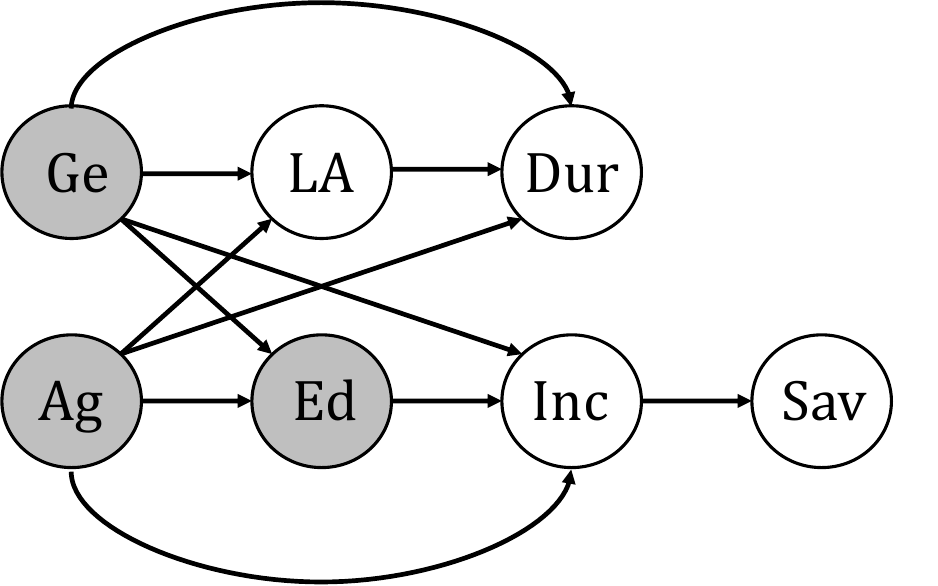}
  \caption{Causal Graph of Loan dataset}
  \label{fig:loan_causal_graph}
\end{figure}

The structural functions of the SCM underlying the \textit{Loan} dataset are, according to \citet{CARMA}, defined as follows:
\begin{enumerate}
    \item $\tilde{f}_{Ge}: Ge = U_{Ge}$ with $U_{Ge} \sim Bernoulli(0.5)$
    \item $\tilde{f}_{Ag}: Ag = -35 +U_{Ag}$ with $U_{Ag} \sim Gamma(10,3.5)$
    \item $\tilde{f}_{Ed}: Ed = -0.5 + \sigma(-1 + 0.5 \cdot Ge + \sigma(0.1 \cdot Ag) + U_{Ed})$ with $U_{Ed} \sim \mathcal{N}(0,0.25)$
    \item $\tilde{f}_{LA}: LA = 0.01 \cdot (Ag -5) \cdot (Ag -5)+ (1-Ge) + U_{LA}$ with $U_{LA} \sim \mathcal{N}(0,4)$
    \item $\tilde{f}_{Dur}: Dur = -1+0.1\cdot Ag + 2 \cdot (1-Ge) + LA + U_{Dur}$ with $U_{Dur}\sim \mathcal{N}(0,9)$
    \item $\tilde{f}_{Inc}: Inc = -4+0.1 \cdot (Ag+35) + 2\cdot Ge + Ge \cdot Ed + U_{Inc}$ with $U_{Inc} \sim \mathcal{N}(0,4)$
    \item $\tilde{f}_{Sav}: Sav = -4 + 1.5 \cdot \mathds{1}\{Inc >0\} \cdot Inc + U_{Sav}$ with $U_{Sav} \sim \mathcal{N}(0,0.25)$
    \item $\tilde{f}_{Y}: Y = \mathds{1}\{y \geq 0\}$ with 
    \\ $y = \sigma(0.3 \cdot (-LA-Dur+Inc+Sav+\alpha \cdot  Inc \cdot Sav))$ 
    \\ and $\alpha = 1$ if $\mathds{1}\{Inc >0\}  \land \mathds{1}\{Sav >0\}$, otherwise $\alpha=-1$ 

\end{enumerate}
In the above equations, $\sigma$ denotes the sigmoid function, 
$\mathcal{N}(a,b)$ denotes the Gaussian distribution with mean $a$ and variance $b$ and $Gamma(a,b)$ denotes the Gamma distribution where $a$ is the concentration/shape, and $b$ is the scale. $Bernoulli(a)$ denotes the Bernoulli distribution with $a$ determining the probability that the random variable takes the value $1$.

\subsubsection{Downstream classifier and causal approximation}
For the downstream automated decision-making model, we consider a $32\times32$ neural network with ReLU activation. We train the classifier with a batch size of 256 for 500 epochs and $10^{-3}$ learning rate (Adam optimizer).

For iCARMA and for estimating log-densities, we approximate the causal relations and the joint data distribution using the state-of-the-art in causal generative modeling, the causal normalizing flows (CNF)~\cite{CausalNF}. For the CNF, we hyperparameter-tuned for different types of normalizing flow layers, the number of hidden layers in each flow layer, etc. After hyperparameter tuning, the best parameters were \emph{masked autoregressive flow}, inner dimension 64 with Elu activation. We used Adam optimizer with $10^{-3}$ learning rate.

\subsubsection{Non-amortized recourse with Oracle}
For the semi-synthetic setup, we consider the causal oracle that uses the true SCM to perform grid search over the actionable space and find optimal recourse recommendations. We set the number of bins for each feature to 25 and create bins around $[-2x_f, 2x_f]$ for some factual feature value $x_f$ while ensuring the feature value lies within the feasible range (minimum and maximum from the training dataset). An increase in the number of bins beyond 25 resulted in prohibitively long runtimes for three to four actionable features.

\subsubsection{Amortized recourse with iCARMA}
\label{Params}
We perform \emph{two separate} hyperparameter tunings, one for considering \emph{ranking-elicited preferences} and another for \emph{score-elicited preferences}. For each preference setup, as mentioned before, we uniformly randomly sample how many and which of the four globally actionable features are individually actionable for each individual. We tuned parameters of the mask and action networks, the different regularizer terms of our loss function, the number of epochs, batch size, and the learning rate for the Adam optimizer. We used Optuna to find the best parameters, balancing the recourse cost and validity.

For rank-based preference, the best parameters were: mask network ($32\times32$), action network ($32\times32\times32$), batch size 128 with 450 epochs, learning rate 0.005. For the loss function, the hinge margin $\beta$ of iCARMA (from original CARMA) was set to 0.013. The $\tau$ of the Gumbel scaling (from CARMA) was set to 0.36. The cost was weighted to 1.1, plausibility regularizer by 0.225, feasibility regularizer by 0.1, and the KL loss of the mask network by 1.5.

For score-based preference, the best parameters were: mask network ($32\times32$), action network ($32\times32$), batch size 128 with 900 epochs, learning rate 0.001. The hinge-margin $\beta$ was 0.004, $\tau$ 0.49, cost was weighted by 1.3, KL by 0.9, with plausibility and feasibility regularizers being scaled by 0.125 and 0.1, respectively.

For each setup, we run iCARMA over 10 random seeds, from 1000 to 10000. For the main paper, we report the performance of the iCARMA model that provided the best validation result and report the variance shown across the deployment-stage population.

\subsection{Real-world Dataset}
\label{apx:real-world-details}
\begin{figure}[th]
    \centering
    \includegraphics[width=0.6\linewidth]{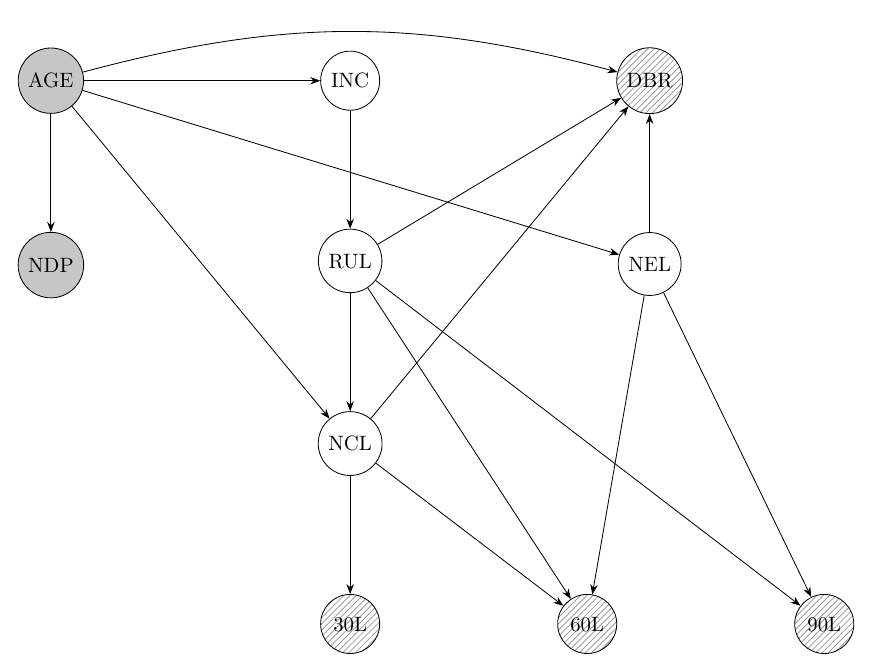}
    \caption{Causal graph for Give Me Some Credit. Solid gray nodes indicate globally immutable, light gray shaded nodes indicate mutable but non-actionable, and white nodes indicate globally actionable nodes.}
    \label{fig:gmsc_graph}
\end{figure}

\subsubsection{Dataset Details}
The causal graph for the Give Me Some Credit dataset is shown in Figure~\ref{fig:gmsc_graph} for the features Age (AGE), Number of Dependents (NDP), Income (INC), Revolving Utilization of Open Lines (RUL), Number of Credit Lines (NCL), Number of Real-Estate Loans (NEL), Debt Ratio (DBR), Number of times 30 days late (30L), Number of times 60 days late (60L), and Number of times 90 days late (90L). 

AGE and NDP are immutable; INC, RUL, NCL, and NEL are globally actionable. DBR, 30L, 60L, and 90L are mutable (from causally upstream interventions) but not directly actionable.
For preprocessing the real-world data, we follow the steps of~\cite{2022personalized_pref_eliction} and their publicly available code\footnote{https://github.com/unitn-sml/pear-personalized-algorithmic-recourse}.
We split the dataset as 0.5 for training, 0.25 for validation, and 0.25 for deployment testing.

\begin{figure}[h]
    \centering
    \includegraphics[width=0.65\linewidth]{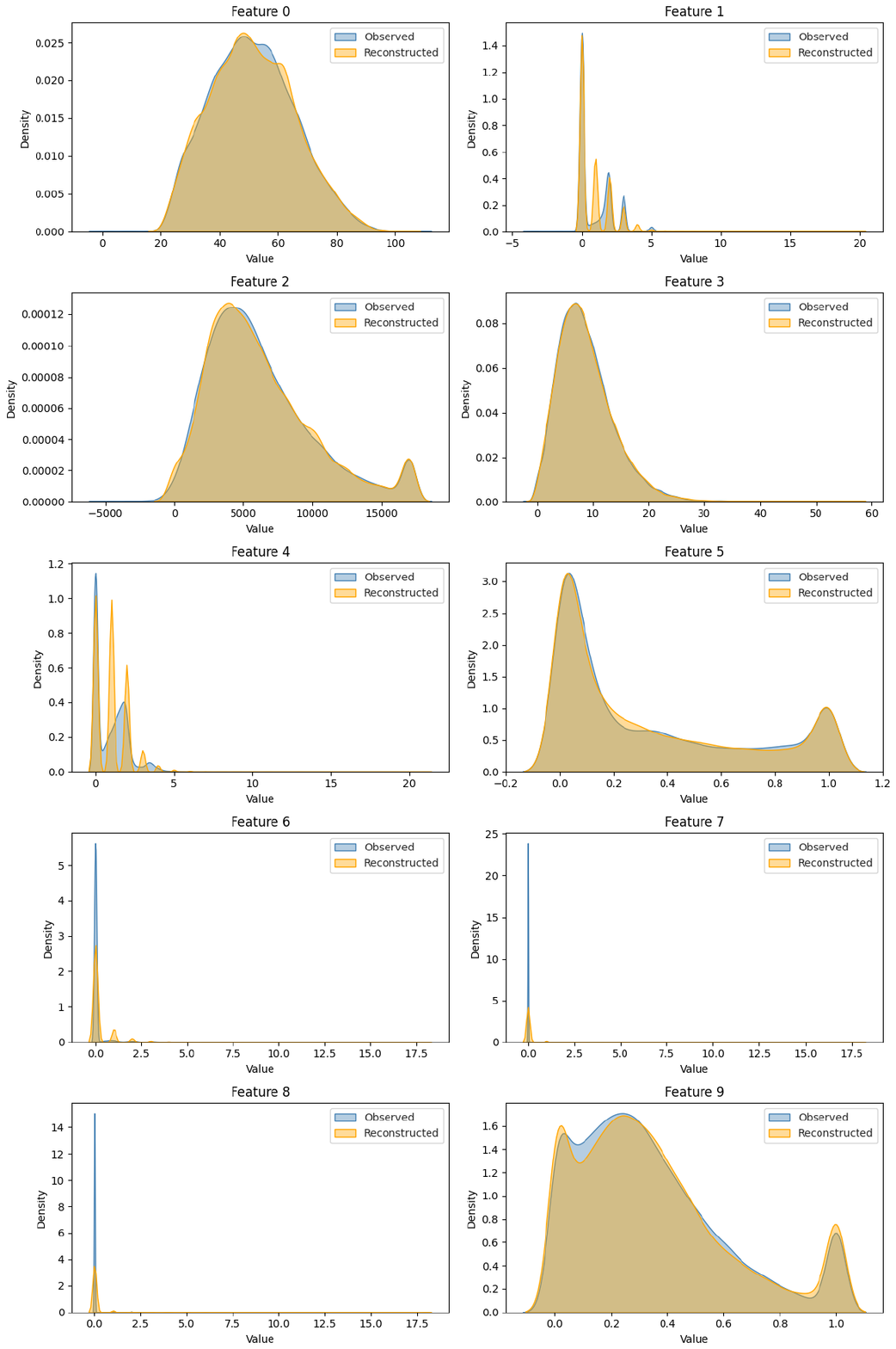}
    \caption{Feature reconstruction for Give Me Some Credit by the best performing CNF model.}
    \label{fig:cnf_gmsc}
\end{figure}

\subsubsection{Downstream classifier and causal approximation}
We use the same architecture for the classifier as we used for the semi-synthetic Loan dataset. However, since the dataset is inherently class-imbalanced, we applied cost-sensitive classification loss to ensure the automated model has reasonable accuracy and default prediction rates as the original data's default rates. After training, the classifier achieves 0.89 accuracy and a default prediction rate of 0.1, matching the original data.

We then hyperparameter-tune to find the best CNF model that approximates the data distribution. The best parameters found for the CNF are: neural spline flow with 32 dimensionality, Elu activation. As we see from Figure~\ref{fig:cnf_gmsc}, the CNF can well-approximate the feature distributions of the original data with the provided causal graph.

\subsubsection{Amortized recourse with iCARMA}
The best hyperparameters obtained for iCARMA were as follows: mask network ($8\times8$), action network ($16\times16\times16\times16$), batch size 256 and 400 epochs, learning rate 0.001, hinge-margin $\beta$ 0.013 and $\tau$ scaling for the Gumbel-softmax as 0.33. Furthermore, we set the weights for cost as 1.1, the KL-loss of the mask network as 1.1, and regularizer weights for plausibility as 0.04 and for feasibility as 0.09.

For each setup, we run iCARMA over 10 random seeds, from 1000 to 10000. For the main paper, we report the performance of the iCARMA model that provided the best validation result and report the variance shown across the deployment-stage population.

\begin{table}[t]
\centering
\caption{Impact on recourse metrics for an increasing number of individually actionable features for the hard-constrained case, assuming \textbf{non-uniform sampling of individually actionable features}: we consider a distribution favouring features high in causal order and a distribution favouring features low in causal order (reverse causal order, rco). For the second case, we additionally compare iCARMA trained with individually actionable features sampled uniformly at random with iCARMA trained with individually actionable features sampled from the distribution biased towards features low in causal order.
\textbf{Oracle knows true SCM and uses grid search with hard constraint for plausibility.}}
\label{tab:exp_1_biased}
\resizebox{0.8\columnwidth}{!}{%
\begin{tabular}{p{2.5 cm}ccccccc}
\toprule
\makecell[l]{bias} & \makecell{number of indivdually \\ actionable features} & method & validity & plausibility & cost \\
\midrule
\multirow{10}{=}{causal order \\ aligned } & \multirow{2}{*}{all (4)} & oracle & 0.959 & 1.0 & $0.057_{\text{\tiny$\pm 0.064$}}$ \\
& & iCARMA & 0.998 & 0.864  &  $0.065_{\text{\tiny$\pm 0.075$}}$ \\
\cmidrule{3-6}
 & \multirow{2}{*}{3} & oracle & 0.821 & 1.0 & $0.066_{\text{\tiny$\pm 0.078$}}$ \\
 & & iCARMA & 0.997 & 0.713  & $0.062_{\text{\tiny$\pm 0.073$}}$ \\
\cmidrule{3-6}
 & \multirow{2}{*}{2} & oracle & 0.597 & 1.0 & $0.072_{\text{\tiny$\pm 0.088$}}$ \\
& & iCARMA & 0.997 & 0.564 & $0.066_{\text{\tiny$\pm 0.09$}}$ \\
\cmidrule{3-6}
& \multirow{2}{*}{1} & oracle & 0.259 & 1.0 & $0.07_{\text{\tiny$\pm 0.087$}}$ \\
& & iCARMA & 0.886 & 0.297 & $0.056_{\text{\tiny$\pm 0.06$}}$ \\
\cmidrule{3-6}
 & \multirow{2}{*}{random} & oracle & 0.656 & 1.0 & $0.062_{\text{\tiny$\pm 0.074$}}$ \\  
 & & iCARMA & 0.971 & 0.611 & $0.063_{\text{\tiny$\pm 0.076$}}$ \\
\hline 
\multirow{15}{=}{reverse causal order \\ aligned } & \multirow{3}{*}{all (4)} & oracle & 0.959 & 1.0 &$0.057_{\text{\tiny$\pm 0.064$}}$ \\ 
& & iCARMA & 1.0 & 0.870 & $0.074_{\text{\tiny$\pm 0.082$}}$ \\
& & iCARMA (rco-trained) & 1.0 & 0.746 &$0.097_{\text{\tiny$\pm 0.098$}}$ \\
\cmidrule{3-6}
 & \multirow{3}{*}{3} & oracle & 0.778 & 1.0 & $0.074_{\text{\tiny$\pm 0.091$}}$ \\
 & & iCARMA & 0.992 & 0.644  & $0.082_{\text{\tiny$\pm 0.09$}}$ \\
 &  & iCARMA (rco-trained) & 0.993 & 0.587 & $0.097_{\text{\tiny$\pm 0.097$}}$ \\
\cmidrule{3-6}
 & \multirow{3}{*}{2} & oracle & 0.467 & 1.0 & $0.085_{\text{\tiny$\pm 0.103$}}$ \\
 & & iCARMA & 0.988 & 0.399 & $0.099_{\text{\tiny$\pm 0.112$}}$ \\
 & & iCARMA (rco-trained) & 0.961 & 0.398 & $0.099_{\text{\tiny$\pm 0.101$}}$ \\
\cmidrule{3-6}
& \multirow{3}{*}{1} & oracle & 0.166 & 1.0 & $0.084_{\text{\tiny$\pm 0.107$}}$ \\
& & iCARMA & 0.798 & 0.194 & $0.099_{\text{\tiny$\pm 0.097$}}$ \\
& & iCARMA (rco-trained) & 0.785 & 0.210 & $0.095_{\text{\tiny$\pm 0.095$}}$ \\
\cmidrule{3-6}
& \multirow{3}{*}{random}& oracle & 0.60 & 1.0 & $0.071_{\text{\tiny$\pm 0.087$}}$ \\
& & iCARMA & 0.950 & 0.548 & $0.089_{\text{\tiny$\pm 0.099$}}$ \\
& & iCARMA (rco-trained) & 0.935 & 0.498 & $0.097_{\text{\tiny$\pm 0.099$}}$ \\
\bottomrule
\end{tabular}
}
\end{table}

\section{Additional Experimental Results}
\label{additional_res}

\begin{table}[t]
\centering
\caption{Ablation of the plausibility constraint in causal-oracle recourse optimization. Removing the plausibility constraint increases validity but produces less distributionally plausible counterfactuals. The effects of individual actionability preferences on other recourse criteria persist even without plausibility constraints.}
\label{tab:oracle_ablate_plaus}
\resizebox{0.65\columnwidth}{!}{%
\begin{tabular}{llllll}
\toprule
distribution & \begin{tabular}[c]{@{}l@{}}individually\\ actionable\end{tabular} & \begin{tabular}[c]{@{}l@{}}plausibility\\ constraint\end{tabular} & validity & plausibility & \begin{tabular}[c]{@{}l@{}}cost\end{tabular} \\
\midrule
\multirow{4}{*}{uniform random} & \multirow{2}{*}{1} & No & 0.876 & 0.25 & $0.135_{\text{\tiny$\pm 0.151$}}$ \\
 &  & Yes & 0.22 & 1.0 &$0.077_{\text{\tiny$\pm 0.096$}}$ \\
 \cmidrule{3-6}
 & \multirow{2}{*}{2} & No & 0.994 & 0.25 &$0.096_{\text{\tiny$\pm 0.133$}}$\\
 &  & Yes & 0.539 & 1.0 & $0.078_{\text{\tiny$\pm 0.096$}}$ \\
 \midrule
\multirow{4}{*}{causal order} & \multirow{2}{*}{1} & No & 0.923 & 0.28 & $0.119_{\text{\tiny$\pm 0.139$}}$ \\
 &  & Yes & 0.259 & 1.0 &$0.07_{\text{\tiny$\pm 0.087$}}$ \\
 \cmidrule{3-6}
 & \multirow{2}{*}{2} & No & 0.997 & 0.37 &$0.078_{\text{\tiny$\pm 0.112$}}$ \\
 &  & Yes & 0.597 & 1.0 & $0.072_{\text{\tiny$\pm 0.088$}}$ \\
 \midrule
\multirow{4}{*}{reverse causal order} & \multirow{2}{*}{1} & No & 0.838 & 0.198 &$0.148_{\text{\tiny$\pm 0.158$}}$ \\
 &  & Yes & 0.166 & 1.0 &$0.084_{\text{\tiny$\pm 0.107$}}$ \\
 \cmidrule{3-6}
 & \multirow{2}{*}{2} & No & 0.997 & 0.29 &$0.103_{\text{\tiny$\pm 0.126$}}$\\
 &  & Yes & 0.467 & 1.0 &$0.085_{\text{\tiny$\pm 0.103$}}$ \\
 \bottomrule
\end{tabular}%
}
\end{table}

\subsection{Analyzing non-uniform random sampling of hard individualized actionability constraints}
\label{apx:add_res_1}

In the main paper, we analyzed the impact of hard individual actionability constraints on recourse metrics when the hard constraints, i.e., which and how many features each individual finds to be individually non-actionable, were uniformly randomly sampled. In this analysis, we explore two other scenarios where individual preferences are sampled using two differently biased distributions. We explore whether our observations transfer to these different scenarios. 

We first consider the case where the likelihood of a feature being preferred increases the higher it is in the causal graph. We refer to this setup as the \emph{causal order aligned}. Second, we consider the scenario where the likelihood of a feature being preferred increases the deeper it is in the causal graph. We refer to this setup as the \emph{reverse causal order aligned}. We report the results in Table~\ref{tab:exp_1_biased}.

\textit{Results.}
We see similar trends regarding the impact of hard individual actionability constraints and recourse metrics as we saw for uniform random sampling performed in the main paper. Hence, regardless of the constraint sampling procedure, constraining more features to be individually non-actionable leads to significant drops in validity and distributional plausibility of the recourse recommendations for both amortized iCARMA and non-amortized Oracle. 

For this specific dataset, we see that when the sampling is reverse causal order aligned, the drop in validity and plausibility is slightly larger. Moreover, we see that the average recourse costs for valid recommendations also remain slightly higher for the reverse ordered case.

\textit{Does training on the same preference distribution help in amortized recourse?}
In the prior analysis, we maintained the same training procedure for the amortized iCARMA, where, during training, the preferences are sampled completely randomly. Hence, the training and testing stage preference distributions \emph{do not align}. Now, for the reverse causal ordered scenario, we additionally train iCARMA using preferences sampled from the same reverse causal ordered distribution. Our results show that knowing the preference distribution at training time \emph{does not necessarily help} in providing better amortized recourse solutions.

\subsection{Ablating the impact of plausibility constraint for individualized actionability}
\label{apx:add_res_2}

 For the semi-synthetic Loan dataset, we considered the causal Oracle to satisfy the hard constraint for plausibility. This is because the Oracle was not using gradient-based solutions. Hence, instead of it being a loss regularized as in iCARMA, it has to be incorporated as a hard constraint. In this analysis, we ablate the impact of the plausibility constraint and the interplay of individual actionability and recourse metrics across the different hard-constraint-sampling strategies. For the specific cases of one or two actionable features, we report the results in Table~\ref{tab:oracle_ablate_plaus}. We restrict to these settings since it was in these cases where we saw the oracle suffer from very low validity.

 \textit{Results.}
From the results, we clearly see that dropping the plausibility constraint helps the oracle achieve very high validity, but at the expense of low distributional plausibility. With more valid recourse recommendations also come more costly recommendations. This trend is observed across different assumed preference sampling scenarios. However, even if we remove the plausibility constraint, we still see similar trends across the different metrics when imposing hard individual constraints. For instance, looking at the causal ordered and reverse causal ordered cases, we clearly see that as we increase the number of features individuals consider actionable from one to two, validity and plausibility increase. Interestingly, in the absence of plausibility constraints, the Oracle's cost also reduces as more features become actionable individually.

\subsection{Individualized Recourse Actionability in Real-World Data}
\label{real_world_res}

The real-world analysis \textit{mirrors the trends} observed in our earlier semi-synthetic experiments. When users face \emph{hard actionability constraints} (which and how many features actionable being randomly sampled), both recourse validity and plausibility drop compared to the non-personalized case, where all four features are individually actionable.

\begin{wrapfigure}[16]{r}{0.4\columnwidth}
\vspace{-2pt}
  \centering
  \includegraphics[width=0.4\columnwidth]{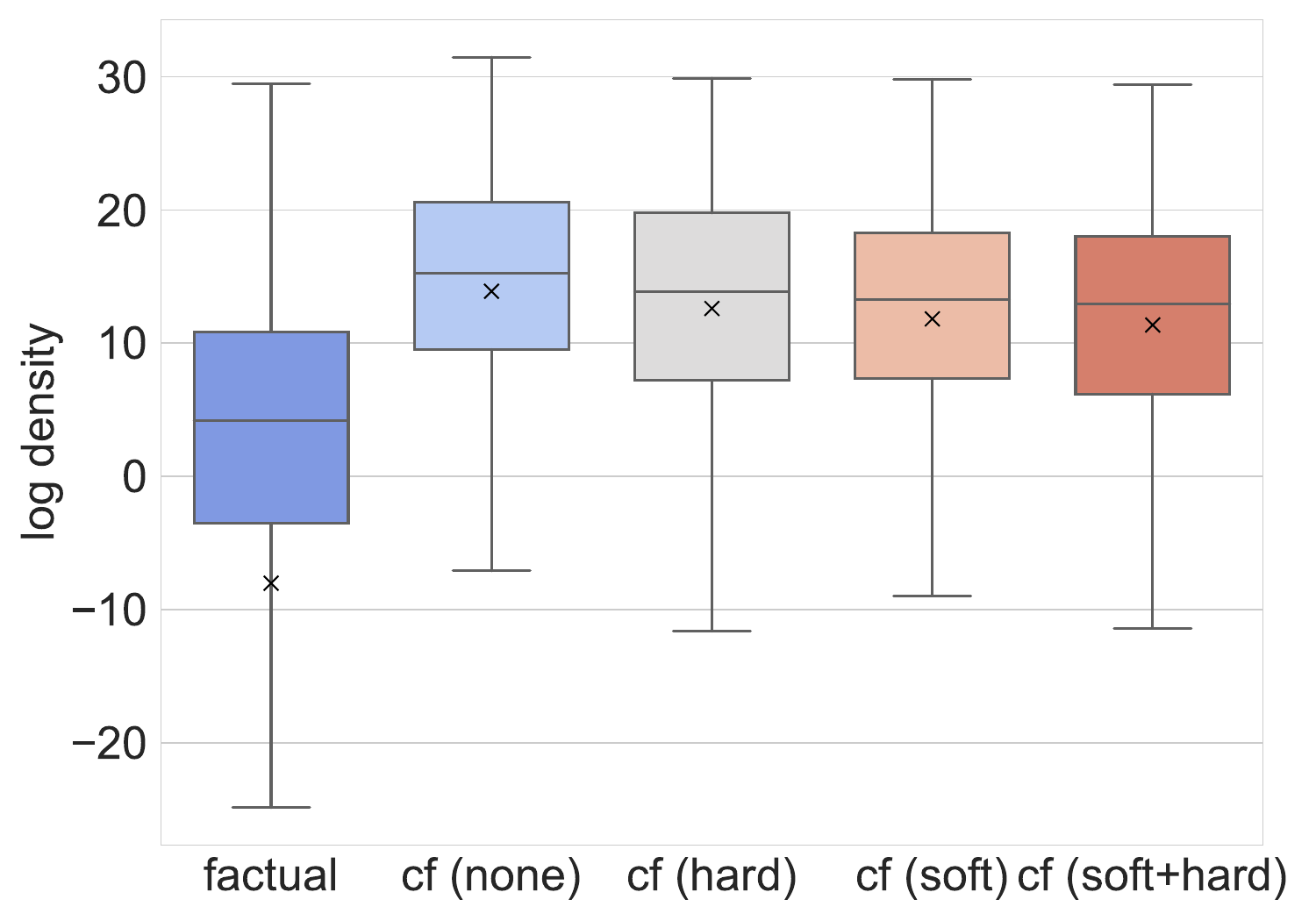}
  \caption{Log-density distribution (iCARMA) for Give Me Some Credit and different personalization setups.}
  \label{fig:gmsc_logdens}
\end{wrapfigure}

\noindent Moving to a \textit{soft} scenario, where all features are actionable, but users provide rankings and a linear cost profile is applied, validity recovers, though at a slightly higher cost and with plausibility still slightly below the non-personalized base case. In the intermediate \emph{hard + soft} setting (users can act on three randomly selected features and provide rankings), the recourse cost decreases. Moreover, introducing individualized cost profiles (linear) reduces the hap metric relative to constant costs, indicating fewer recommendations that target users' least preferred actionable features.

\noindent To better analyze the distributional plausibility effects, we plot the log-densities of the original factuals and the recourse counterfactuals from the different personalization scenarios in Figure~\ref{fig:gmsc_logdens}. Incorporating hard individual actionability constraints reduces the likelihood of generated recourse counterfactuals and increases the spread towards less-likely regions of the data space. In contrast, only having soft constraints results in lower spread and higher log-density values of the recommended counterfactuals.

\noindent \textbf{Takeaway:} This real-world study \textit{reiterates our earlier findings}: satisfying hard individual actionability constraints substantially reduces recourse validity and distributional plausibility. In contrast, having primarily soft individualized constraints has a less severe impact on validity, cost, and plausibility on deployment.

\end{document}